\documentclass[sigconf]{acmart}

\usepackage{lipsum}
\usepackage{graphicx}

\usepackage{caption}
\usepackage{subcaption}

\newcommand{\st}{\scriptscriptstyle}
\usepackage[skins]{tcolorbox}
\newtcolorbox{mybox}[2][]{%
  attach boxed title to top center
               = {yshift=-11pt},
  colframe     =black,
  colbacktitle = black,
  title        = #2,#1,
  enhanced,
}
\newcommand{\et}{\textit{et al}\xspace}

\newcommand{\adv}{\ensuremath{\mathcal{A}}\xspace}

\newcommand{\feat}{\ensuremath{\text{feat}}\xspace}

\usepackage{xspace}
\usepackage{color, colortbl}
\usepackage{enumitem}
\usepackage{amsfonts}
\usepackage{algorithm}
\usepackage[noend]{algpseudocode}

\usepackage{bm}

\usepackage{multirow}
\usepackage{lipsum}%arbitrary text

\newcommand{\fsp}{\ensuremath{\text{FSP}}\xspace}

\newcommand{\aut}{\ensuremath{\text{Srv}}\xspace}

\newcommand{\clr}{blue!15}

\newcommand{\bank}{\ensuremath{\text{C}}\xspace}

\newcommand{\usr}{\ensuremath{\text{U}}\xspace}

\newcommand{\susr}{\ensuremath{{u}}\xspace}

\newcommand{\fea}{\ensuremath{{t}}\xspace}

\newcommand{\feas}{\ensuremath{{T}}\xspace}

\newcommand{\fc}{\ensuremath{\text{FC}}\xspace}

\newcommand{\prt}{\ensuremath{\text{C}}\xspace}

\newcommand{\starlit}{\ensuremath{\textit{Starlit}}\xspace}
\newcommand{\flower}{\ensuremath{\text{Flower}}\xspace}

\newcommand{\deltap}{d_p} % function to measure privacy (attacker's error when estimating v as \hat{v})
\newcommand{\set}[1]{\mathcal{#1}}

\newcommand{\func}{\ensuremath{\mathcal{F}}\xspace}

\newcommand{\leak}{\ensuremath{\mathcal{L}}\xspace}

\newcommand{\leakW}{\ensuremath{\mathcal{W}}\xspace}

\newcommand{\cel}{\textit{Celestial}\xspace}
\newcommand{\empt}{\ensuremath{\bot}\xspace}

\newcommand{\prm}{\ensuremath{prm}\xspace}

\usepackage{url}
\usepackage{mathtools}

\AtBeginDocument{%
  \providecommand\BibTeX{{%
    \normalfont B\kern-0.5em{\scshape i\kern-0.25em b}\kern-0.8em\TeX}}}

\setcopyright{none}
\renewcommand\footnotetextcopyrightpermission[1]{}



\begin{document}

\title{Starlit: Privacy-Preserving  Federated Learning to\\ Enhance Financial Fraud Detection}

%\thanks{This is a footnote to the title.}
%tarlit: Privacy-Preserving Federated Learning to Enhance Financial Fraud Detection

\author{Aydin Abadi}
\authornote{aydin.abadi@ucl.ac.uk}
\affiliation{%
  \institution{University College London}
  \country{}}
%\email{aydin.abadi@ucl.ac.uk}

\author{Bradley Doyle}
\affiliation{%
  \institution{Privitar}
  \country{}}

\author{Francesco Gini}
\affiliation{%
  \institution{Privitar}
  \country{}}

\author{Kieron Guinamard}
\affiliation{%
  \institution{Privitar}
  \country{}}

\author{Sasi Kumar Murakonda}
 \authornote{sasi.murakonda@privitar.com}
\affiliation{%
  \institution{Privitar}
  \country{}}

  \author{Jack Liddell}
\affiliation{%
  \institution{Privitar}
  \country{}}

    \author{Paul Mellor}
\affiliation{%
  \institution{Privitar}
  \country{}}

\author{Steven J. Murdoch}
\authornote{s.murdoch@ucl.ac.uk}
\affiliation{%
  \institution{University College London}
  \country{}
  }

      \author{Mohammad Naseri}
      \authornote{mohammad@flower.dev}
\affiliation{%
  \institution{Flower Labs and\\ University College London}
  \country{}}

    \author{Hector Page}
\affiliation{%
  \institution{Privitar}
  \country{}}
  
    \author{George Theodorakopoulos}
          \authornote{theodorakopoulosg@cardiff.ac.uk}
\affiliation{%
  \institution{Cardiff University}
  \country{}}
  
      \author{Suzanne Weller}
       \authornote{suzanne.weller@privitar.com}
\affiliation{%
  \institution{Privitar}
  \country{}}

%\author{}
%%\institute{}
%\date{}

\begin{abstract}

Federated Learning (FL) is a data-minimization approach enabling collaborative model training across diverse clients with local data, avoiding direct data exchange. 
However, state-of-the-art FL solutions to identify fraudulent financial transactions exhibit a subset of the following
limitations. They (1) lack a formal security definition and proof, (2) assume prior freezing of suspicious customers’ accounts by financial institutions (limiting the solutions' adoption), (3) scale poorly, involving either  $O(n^{\st 2})$ computationally expensive modular exponentiation (where $n$ is the total number of financial institutions) or highly inefficient fully homomorphic encryption, (4)  assume the parties have already completed the identity alignment phase, hence excluding it from the implementation, performance evaluation, and security analysis, and (5) struggle to resist clients' dropouts. 
This work introduces \starlit, a novel \textit{scalable} privacy-preserving FL mechanism that overcomes these limitations.\footnote{Our solution, \starlit, has been awarded the joint first prize of UK--US Privacy-Enhancing Technologies Challenge Prize \cite{PETs2023}. \starlit has been acknowledged by both the White House and UK Government websites \cite{the-white-house,the-UK-government}.} It has various applications, such as enhancing financial fraud detection, mitigating terrorism, and enhancing digital health. We implemented \starlit and conducted a thorough performance analysis using synthetic data from a key player in global financial transactions. The evaluation indicates \starlit's scalability, efficiency, and accuracy.  
\end{abstract}

\pagestyle{plain}
% \setstretch{1.5}

\maketitle{}

% \input{phase2_plan.tex}
% \newpage
% !TEX root =main.tex

%\vspace{-1mm}
\section{Introduction}\label{sec::intro}
%\vspace{-.4mm}

Sharing data is crucial in dealing with crime. Collaborative data analysis among law enforcement agencies and relevant stakeholders can significantly enhance crime prevention, investigation, and overall public safety. 
For instance, in the United Kingdom, Cifas, a non-profit fraud database, and fraud prevention organization that promotes data sharing among its members, reported that its members detected and reported over 350,000 cases of fraud in 2019. This collective effort prevented fraudulent activities amounting to £1.5 billion \cite{white2021financial}.  
The National Data Sharing Guidance, developed by the UK Home Office and Ministry of Justice in 2023, further underscores the importance of data sharing in dealing with crime \cite{data-sharing-UK-guidance}.

Typically, inputs for collaborative data analysis come from different parties, each of which may have concerns about the privacy of their data. Federated Learning (FL) \cite{YangLCT19} and secure Multi-party Computation (MPC) \cite{Yao82b}, along with their combination, are examples of mechanisms that allow parties to collaboratively analyze shared data while maintaining the privacy of their input data.%\footnote{The primary difference between FL and MPC is that in the former, parties jointly compute a global \textit{model} while in the latter, they jointly compute a certain \textit{function} of the inputs.} 

FL is a machine learning framework where multiple parties collaboratively build machine learning models without revealing their sensitive input to their counterparts \cite{YangLCT19,McMahanMRA16}. Vertical Federated Learning (VFL) is a vital variant of FL, with various applications, e.g., in dealing with crime \cite{abs-2310-04546} and healthcare \cite{NguyenPPDSLDH23}. VFL refers to the FL setting
where datasets distributed among different parties (e.g., banks) have some intersection concerning users (e.g., have certain customers' names in common)  while holding different features, e.g., customers' names, addresses, and how they are perceived by a financial institution. 
Horizontal Federated Learning (HFL) is another important variant of FL where participants share the same feature space while holding different users, e.g., customers' attributes are the same, but different banks may have different customers.

Advanced privacy-preserving FL-based solutions aiming to detect anomalies and deal with financial fraud may face a new challenge. In this setting, datasets for financial transactions might be partitioned both vertically and horizontally. 
For instance, a third-party Financial Service Provider (\fsp) may have details of financial transactions including customers' names, and involved banks, while each \fsp's partner bank may have some details/features of a subset of these customers. Thus, existing solutions for VFL or HFL cannot be directly applied to deal with this challenge.

% Concurrently with this work, researchers have proposed VFL-based solutions in \cite{abs-2310-04546,abs-2305-11236} to deal with financial crime. Nevertheless, as discussed in Section \ref{sec:sec:Distributed-Approaches}, they exhibit a subset of the following limitations: (1) they lack a formal security definition and proof, (2) they assume that a financial institution has already frozen suspicious customers' accounts (which limits the solutions' adoption), (3) they involve $O(n^{\st 2})$ computationally expensive modular exponentiation where $n$ is the total number of financial institutions, (4) they assume the parties have already completed the identity alignment phase, hence excluding it from the implementation, performance evaluation, and security analysis, and (5) they fail to terminate successfully even if only one of the clients neglects to transmit its message, i.e., cannot resist clients' dropouts. 

%\vspace{-2mm}
\subsection{Our Contributions}

In this work, we introduce \starlit, a pioneering scalable privacy-preserving federated
learning mechanism that can help enhance financial fraud detection. 
By devising and utilizing \starlit in the context of financial fraud, we address all limitations of the state-of-the-art FL-based mechanisms, proposed in \cite{abs-2310-04546,abs-2305-11236,abs-2310-19304}. Specifically, we (1) formally define and prove \starlit's security (in the simulation-based paradigm), (2) do not place any assumption on how suspicious accounts of customers are treated by their financial institutions, (3) make \starlit scale linearly with the number of participants (i.e., its overhead is $O(n)$) while refraining from using fully homomorphic encryption, (4) include all phases of \starlit in the implementation, performance evaluation, and security analysis, and (5) make \starlit resilient against dropouts of clients.%, e.g.,  financial institutions. 

\starlit offers two compelling properties not found in existing VFL schemes. These include the ability to securely:
\begin{itemize}[leftmargin=4.8mm]
\item Identify discrepancies among the values of shared features in common users between distinct clients' datasets. For instance, in the context of banking, \fsp and a bank can detect if a certain customer provides a different home address to each.%, without learning anything else.  

\item Aggregate common features in shared users among different clients' datasets, even when these features have varying values. For instance, this feature will enhance   \fsp's data by reflecting whether  \fsp and multiple banks consider a certain customer suspicious, according to the value of a flag independently allocated by each bank to that customer's account.

\end{itemize}

We have implemented \starlit and evaluated its performance using synthetic data which comprises about four million rows. This synthetic data was provided by a major organization globally handling financial transactions.  
\starlit stands out as the first solution that simultaneously provides the features mentioned above. We identify several potential applications for \starlit, including mitigating terrorism, enhancing digital health, and aiding in the detection of benefit fraud (see Section \ref{sec::other-usecases}).

To develop \starlit, we use a combination of several tools and techniques, such as SecureBoost (for VFL), Private Set intersection (for identity alignment and finding discrepancies among different entities' information), and Differential Privacy to preserve the privacy of accounts' flags (that indicate whether an account is deemed suspicious). Moreover, based on our observation that each dataset's sample (or row), such as a financial transaction, can be accompanied by a random identifier, we allow a third-party feature collector to efficiently aggregate clients' flags without being able to associate the flags values with a specific feature, e.g., customer's name.

\

\noindent\textbf{Summary of our Contributions.} In this work, we:
\begin{itemize}[leftmargin=4mm,label=$\bullet$]
    \item Introduce \starlit, a novel scalable privacy-preserving federated learning mechanism, with various real-world applications.%, e.g., in enhancing financial fraud detection, mitigating terrorism, and digital healthcare. 

    \item Formally define and prove \starlit's security using the simulation-based paradigm.

    \item Implement \starlit and conduct a comprehensive evaluation of its performance.

    %\item Demonstrate several real-world applications of \starlit.
\end{itemize}

% We formally define the security of  \starlit in the standard simulation-based paradigm and prove its security. We have implemented \starlit and comprehensively evaluated its performance using synthetic data provided by one of the major players in accommodating financial transactions. \starlit exhibits a high level of scalability as its overheads grow linearly with the number of financial institutions that do not need to interact with each other or even know the existence of other participants.  

% ** We present the first privacy-preserving data-sharing system developed specifically for financial transactions.

% ** We have implemented the system and thoroughly analysed its accuracy and cost. 

%\vspace{-2mm}

\subsection{Primary Goals and Setting}\label{sec::problem-statement}

This paper focuses on a real-world scenario in which a server, denoted by \aut, wants to train a machine-learning model to detect anomalies using its data, and complementary data held by different clients $C=\{\prt_{\st 1},..., \prt_{\st m}\}$. 
For instance, \aut can be a Financial Service Provider (\fsp) such as SWIFT\footnote{\url{https://www.swift.com}}, Visa\footnote{\url{https://www.visa.co.uk/about-visa.html}}, PayPal\footnote{\url{https://www.paypal.com/uk/home}}, CHIPS\footnote{\url{https://www.theclearinghouse.org/payment-systems/chips}}, and SEPA\footnote{\url{https://finance.ec.europa.eu/consumer-finance-and-payments/payment-services/single-euro-payments-area-sepa_en}}—facilitating financial transactions and payments between various clients in set $C$, such as banks, eBay, and Amazon—that aims to detect anomalous transactions. 

In this setting, \aut may maintain a database of samples/rows between interacting clients, but it does not possess all the details about the users included in each sample. For instance, in the context of financial transactions, \fsp holds a dataset containing samples (i.e., transactions) between the ordering account held by bank $\prt_{\st i}$ and the beneficiary account held by bank $\prt_{\st j}$. 

Each sample may contain a customer's name, the amount sent, home address, and information about $\prt_{\st i}$, and $\prt_{\st j}$. Each client in $C$ maintains a dataset containing certain customers' account information, including customers' details, their transaction history, and even local assessments of their known financial activities. However, each $\prt_{\st j}$ may not hold all users (e.g., customers) that \aut is interested. 

While \aut is capable of training a model to detect anomalous transactions using its data, it could enhance the analytics by considering the complementary data held by other clients concerning the entities involved in the transactions. 
The ultimate goal is to enable \aut to collaborate with other clients to develop a model that is significantly better than the one developed on \aut's data alone, e.g., to detect suspicious transactions and ultimately to deal with financial fraud. 

However, a mechanism that offers the above feature must satisfy vital security and system constraints; namely, (i) the privacy of clients' data should be preserved from their counterparts, and (ii) the solution must be efficient for real-world use cases. 
%
% However, it is required that no sensitive data (for instance about ordinary customers) is shared between \fsp and the individual entities, and the solution should be efficient enough to be used in practice. 
%
The aforementioned setting is an example of FL on vertically and horizontally partitioned data in which each \aut's transaction is associated with a sender  $\prt_{\st i}$ (e.g., ordering bank), and receiver $\prt_{\st j}$, e.g., beneficiary bank. 
Our solution will enhance \aut's dataset with two primary types of features using the datasets of $\prt_{\st i}$ and $\prt_{\st j}$:

\begin{itemize}[leftmargin=4.2mm]

   \item[$\bullet$] {\bf Discrepancy Feature}: This will enhance  \aut's data by reflecting whether there is a discrepancy between (i) the (value of the) feature, such as a customer's name and address, it holds about a certain user $\usr$ under investigation and (ii) the feature held by sending client $\prt_{\st i}$ and receiving client $\prt_{\st j}$ about the same user. For each user, this feature is represented by a pair of binary values $(b_{\st \susr, i}, b_{\st \susr,j})$, where  $b_{\st \susr, i}$  and $b_{\st \susr,j}$ represents whether the information that \aut holds matches the one held by the sending and receiving clients respectively.

    % \item[$\bullet$] {\bf Discrepancy Feature}: This feature will enhance   \aut's data by reflecting whether there is a discrepancy between (i) the information (e.g., name and address) it holds about a certain customer $c$ under investigation and (ii) the information held by the ordering entity $\prt_{\st i}$ and held by the beneficiary entity $\prt_{\st j}$ about the same customer. For each customer, this feature is represented by a pair of binary values $(b_{\st c,i}, b_{\st c,j})$, where  $b_{\st c,i}$  and $b_{\st c,j}$ represents whether the information that \fsp holds matches the one held by the ordering and beneficiary entities respectively. 

    \item[$\bullet$] {\bf Sample's Flag Feature}: This will enhance   \aut's data by reflecting whether  \aut and a client have the same view of a certain user, e.g., a customer is suspicious. This feature is based on a pair of binary private flags for a certain user, where one flag is held by the sending client and the other one is held by the receiving client. In the context of banking, banks often allocate flags to each customer's account for internal use. The value of this flag is set based on the user's transaction history and determines whether the bank considers the account holder suspicious.

    % is taken directly from the dataset of the ordering and beneficiary accounts respectively - just two binary values (one representing whether the flag value is zero or not for the ordering account and the other for the beneficiary account)

\end{itemize}

To preserve the privacy of the participating parties' data (e.g., data of non-suspicious customers held by banks) while aligning \aut's dataset with the features above, we rely on a set of privacy-enhancing techniques, such as Private Set Intersection (PSI) and Differential Privacy (DP). 
Briefly, to enable \aut to find out whether the data it holds about a certain (suspicious) user matches the one held by a client, we use PSI. Furthermore, to enhance \aut's data with the flag feature, each client uses local DP to add noise to their flags and sends the noisy flags to a third-party flag collector which feeds them to the model training phase.

% Our system ensures that the data at the banks (such as names and addresses) is never disclosed in plaintext to \fsp. Furthermore, it ensures that banks only share minimal information that is necessary for improving anomaly detection models. Moreover, for a desired level of improvement in accuracy, our framework finds and applies the mechanism that provides the maximum protection to flag values at the banks.

%\newpage % only for ease when writing, to be removed

% !TEX root =main.tex

%\vspace{-1mm}
\section{Related Work}

In this section, we briefly discuss the privacy-preserving FL-based approaches used to deal with fraudulent transactions. We refer readers to Appendix \ref{sec::survey} for a survey of related work. 
Lv \et. \cite{LvCZYMW21} introduced an approach to identify black market fraud accounts before fraudulent transactions occur. It aims to guarantee the safety of funds when users transfer funds to black market accounts, enabling the financial industry to utilize multi-party data more efficiently.  
It involves data provided by financial and social enterprises. The approach utilizes \textit{insecure} hash-based PSI for identity alignment.

This scheme differs from \starlit in a couple of ways: (i) \starlit operates in a multi-party setting, where various clients contribute their data, in contrast to the aforementioned scheme, which has been designed for only two parties, and (ii) \starlit deals with the data partitioned both horizontally and vertically, whereas the above scheme focuses only on vertically partitioned data.

% Recently, to combat the global challenge of organized crime, such as money laundering, terrorist financing, and human trafficking,  the UK and US governments launched a set of prize challenges \cite{PETs2023}. This competition encouraged innovators to develop technical solutions to identify suspicious bank account holders while preserving the privacy of honest account holders by relying on FML and cryptography approaches. This underscores the importance the distributed privacy-preserving financial data analytics for governments. 

Recently, Arora \et.  \cite{abs-2310-04546} introduced an approach that relies on oblivious transfer, secret sharing, DP, and multi-layer perception. The authors have implemented the solution and conducted a thorough analysis of its performance.

\noindent\underline{\textit{\starlit versus the Scheme of Arora \et.}}  
The latter assumes that the ordering bank never allows a customer with a dubious account to initiate transactions but allows the same account to receive money. In simpler terms, this scheme exclusively addresses frozen accounts, restricting its applicability. This setting will exempt the ordering bank from participating in MPC, enhancing the efficiency of the solution. 

In the real world, users' accounts might be deemed suspicious (though not frozen), yet they can still conduct financial transactions within their bank.  The bank may handle such accounts more cautiously than other non-suspicious accounts. In contrast, \starlit (when applied to financial transactions context) does not place any assumption on how a bank treats a suspicious account.

Furthermore, unlike the scheme proposed in \cite{abs-2310-04546}, which depends on an ad-hoc approach to preserve data privacy during training, our solution, \starlit, employs SecureBoost—a well-known scheme extensively utilized and analyzed in the literature. 
Thus, compared to the scheme in \cite{abs-2310-04546}, Starlit considers a more generic scenario and relies on a more established scheme for VFL.

Recently, another approach has been developed by Qiu \et. \cite{abs-2305-11236}. It uses neural networks and shares the same objective as the one by Arora \et. However, it strives for computational efficiency primarily through the use of symmetric key primitives. The scheme incorporates the elliptic-curve Diffie-Hellman key exchange and one-time pads to secure exchanged messages during the model training phase. This scheme has also been implemented and subjected to performance evaluation.

\noindent\underline{\textit{\starlit versus the Scheme of Qiu \et.}}  The latter scheme requires each client (e.g., bank) to possess knowledge of the public key of every other client and compute a secret key for each through the elliptic-curve Diffie-Hellman key exchange scheme. Consequently, this approach imposes $O(n)$ modular exponentiation on each client, resulting in the protocol having a complexity of  $O(n^{\st 2})$, where $n$ represents the total number of clients. In contrast, in \starlit, each client's complexity is independent of the total number of clients and each client does not need to know any information about other participating clients.  
Moreover, the scheme proposed in \cite{abs-2305-11236} assumes the parties have already performed the identity alignment phase, therefore, the implementation, performance evaluation, and security analysis exclude the identity alignment phase.

Furthermore, the scheme in \cite{abs-2305-11236} fails to terminate successfully even if only one of the clients neglects to transmit its message. In this scheme, each client, utilizing the agreed-upon key with every other client, masks its outgoing message with a vector of pseudorandom blinding factors. The expectation is that the remaining clients will mask their outgoing messages with the additive inverses of these blinding factors. These blinding factors are generated such that, when all outgoing messages are aggregated, the blinding factors cancel each other out. 

Nevertheless, if one client fails to send its masked message, the aggregated messages of the other clients will still contain blinding factors, hindering the training on correct inputs.  
%
% The solution proposed in \cite{BonawitzIKMMPRS17}, based on threshold secret sharing, can address this issue. However, incorporating such a patch would introduce additional computation and communication overheads.
In contrast, \starlit does not encounter this limitation. This is because the message sent by each client is independent of the messages transmitted by the other clients.

Kadhe \et. \cite{abs-2310-19304} proposed an anomaly detection scheme, that uses fully homomorphic encryption (computationally expensive), DP, and secure multi-party computation. The authors have also implemented their solution and analyzed its performance.

\noindent\underline{\textit{\starlit versus the Scheme of Kadhe \et.}} The latter heavily relies on fully homomorphic encryption. In this scheme, all parties need to perform fully homomorphic operations. This will ultimately affect both the scalability and efficiency of this scheme. In contrast, \starlit does not use any fully homomorphic scheme.

All of the above solutions share another shortcoming, they lack formal security definitions and proofs of the proposed systems.

% !TEX root =Feather.tex

%\vspace{-2mm}
\section{Informal Threat Model}\label{sec::threat-model}
% \sasi{I am editing this section to portray our story as one set of solutions with LDP and the other with SecureBoost+ optional DP). This will be followed by empirical results/trade-off analysis for each. And then technical details of all the parts...}

\starlit involves three types of parties: 

\begin{itemize}[leftmargin=4.5mm]

\item[$\bullet$] \underline{Server (\aut)}. 
It wants to train a model to detect anomalies using its data, and complementary data held by different
clients. The data \aut maintains is partitioned vertically and horizontally across different clients. Each sample in the data includes various features, e.g., a user's name, sender client, and receiver client.

% It facilitates financial transactions and payments between various entities. \fsp often observes and maintains the financial messages, particularly transactions exchanged between the bank that sends a transaction and the bank that receives the transaction.

% \item[$\bullet$] \underline{Financial Service Provider (\fsp)}. It facilitates financial transactions and payments between various entities. \fsp often observes and maintains the financial messages, particularly transactions exchanged between the bank that sends a transaction and the bank that receives the transaction.

\item[$\bullet$] \underline{Clients  ($\bank_{\st 1},..., \bank_{\st n}$)}. They are different clients (e.g., nodes, devices, or organizations) that contribute to FL by providing local complementary data to the training process.

% regular banks that provide financial services to their customers. It is possible that multiple banks share a common set of customers. The beneficiary bank refers to a bank to whom the payment is to be made. The ordering bank, on the other hand, is the bank that a customer uses to issue a letter of credit. This bank is often referred to as the issuing bank because it issues the letter of credit on behalf of its client.

\item[$\bullet$] \underline{Flag Collector (\fc)}. It is a third-party helper that aggregates some of the features held by different clients. \fc is involved in \starlit to enhance the system's scalability.

\end{itemize}

% \fsp, Bank Clients, and Flag Collector (a separate client or an independent entity being run at one of other two clients). 

We assume that all the participants are honest but curious (a.k.a. passive adversaries), as it is formally defined in \cite{DBLP:books/cu/Goldreich2004}. Hence, they follow the protocol's description. But, they try to learn other parties' private information. 
We consider it a privacy violation if the information about one party is learned by its counterpart during the model training (including pre-processing). We assume that parties communicate with each other through secure
channels.

\section{Preliminaries}\label{sec::preliminaries}

\subsection{Notations and Assumptions}
Table \ref{table:notation-table} summarizes the notations used in this paper. Let $\mathcal{G}$ be a  multi-output function, $\mathcal{G}(inp)\rightarrow (outp_{\st 1},..., outp_{\st n})$. Then, by  $\mathcal{G}_{\st i}(inp)$ we refer to the $i$-th output of $\mathcal{G}(inp)$, i.e.,  $outp_{\st i}$.

% !TEX root =main.tex

%\break

%\section{Notations}\label{sec:notation-table}
%
% We summarise our notations in Table \ref{table:notation-table}.

\begin{table}[ht]
\begin{scriptsize}
\begin{center}
\footnotesize{
%\vspace{-3mm}
\caption{ \small{Notation table}.}\label{commu-breakdown-party} 
%\vspace{-2.5mm}
\renewcommand{\arraystretch}{1.1}
\scalebox{0.97}{
% 1st table
\begin{tabular}{|c|c|c|c|c|c|c|c|c|c|c|c|c|c|} 

\hline 

\cellcolor{\clr} \scriptsize \textbf{Symbol}&\cellcolor{\clr} \scriptsize \textbf{Description}  \\
    \hline
    
     \hline

%Generic
%\multirow{29}{*}{\rotatebox[origin=c]{90}{\scriptsize \textbf{Generic}}}

 \cellcolor{gray!20}\scriptsize \aut &\cellcolor{gray!20}\scriptsize Server \\

 \cellcolor{white!20}\scriptsize \fsp &\cellcolor{white!20}\scriptsize Financial Service Provider \\

 \cellcolor{gray!20}\scriptsize \fc &\cellcolor{gray!20}\scriptsize Feature Collector\\ 

 \cellcolor{white!20}\scriptsize $\bank_{\st i}$ &\cellcolor{white!20}\scriptsize A client or bank\\ 

%\bank_{\st 1}

 \cellcolor{gray!20}\scriptsize   &\cellcolor{gray!20}\scriptsize  Parameter that quantifies the privacy guarantee\\ 

\multirow{-2}{*}{\cellcolor{gray!20}\scriptsize $\epsilon$}  &\cellcolor{gray!20}\scriptsize  provided by a differentially private mechanism.\\

 \cellcolor{white!20}\scriptsize PSI &\cellcolor{white!20}\scriptsize Private Set Intersection\\

 \cellcolor{gray!20}\scriptsize DP &\cellcolor{gray!20}\scriptsize Differential Privacy\\

 \cellcolor{white!20}\scriptsize ML &\cellcolor{white!20}\scriptsize Machine Learning\\ 

  \cellcolor{gray!20}\scriptsize FL &\cellcolor{gray!20}\scriptsize Federated Learning\\ 

 \cellcolor{white!20}\scriptsize VFL &\cellcolor{white!20}\scriptsize Vertical Federated Learning\\ 

 \cellcolor{gray!20}\scriptsize RR &\cellcolor{gray!20}\scriptsize Randomized Response\\

\cellcolor{white!20}\scriptsize AUPRC  &\cellcolor{white!20}\scriptsize  Area Under the Precision-Recall Curve\\ 

\cellcolor{gray!20}\scriptsize GOSS  &\cellcolor{gray!20}\scriptsize  Gradient-based One Side Sampling\\ 

\cellcolor{white!20}\scriptsize H  &\cellcolor{white!20}\scriptsize  Hour\\

  \cellcolor{gray!20}\scriptsize $Pr$ &\cellcolor{gray!20}\scriptsize Probability \\

    \cellcolor{white!20}\scriptsize $S_{\st i}$ &\cellcolor{white!20}\scriptsize A private set\\ 
 
\cellcolor{gray!20}\scriptsize$\set{V}$ &\cellcolor{gray!20}\scriptsize Set of flag values  \\

 \cellcolor{white!20}\scriptsize$\pi(v), v \in \set{V}$&\cellcolor{white!20}\scriptsize Prior probability of value $v$ (\fsp's prior knowledge)\\

\cellcolor{gray!20}\scriptsize$\deltap(\hat{v}, v) : \set{V} \times \set{V} \to \set{R}$ &\cellcolor{gray!20}\scriptsize  Privacy metric (attacker's error when estimating $v$ as $\hat{v}$)\\

\cellcolor{white!20}\scriptsize $f(v' | v) : \set{V} \times \set{V} \to a \in \{0,1\}$  &\cellcolor{white!20}\scriptsize  Privacy mechanism\\   

\cellcolor{gray!20}\scriptsize $||$  &\cellcolor{gray!20}\scriptsize  Concatenation\\

\cellcolor{white!20}\scriptsize $\mathcal{L}$  &\cellcolor{white!20}\scriptsize  Leakage function of \cel and \starlit\\

\cellcolor{gray!20}\scriptsize $\mathcal{L}_{\st 1}$  &\cellcolor{gray!20}\scriptsize  \fsp–side leakage in \starlit \\  

\cellcolor{white!20}\scriptsize $\mathcal{L}_{\st 2}$  &\cellcolor{white!20}\scriptsize  \fc–side leakage in \starlit \\

\cellcolor{gray!20}\scriptsize $\mathcal{L}_{\st i+2}$  &\cellcolor{gray!20}\scriptsize  $\bank_{\st i}$–side leakage in \starlit \\

\cellcolor{white!20}\scriptsize $\mathcal{W}$  &\cellcolor{white!20}\scriptsize  Leakage function in (V)ML \\

\cellcolor{gray!20}\scriptsize $\mathcal{F}$  &\cellcolor{gray!20}\scriptsize  Functionality of \cel \\

\cellcolor{white!20}\scriptsize $\prm_{\st i}$  &\cellcolor{white!20}\scriptsize  Input parameter of a party to (V)FL \\

\cellcolor{gray!20}\scriptsize $|S|$  &\cellcolor{gray!20}\scriptsize  Size of set or database $S$ \\

         \hline

\end{tabular}\label{table:notation-table}

}}
\end{center}
\end{scriptsize}
%\vspace{-3mm}
\end{table}

%%%%%%%%%%%%%%%%%%%%%%%%%%%%%%%%%%%%%%%%%%%

% !TEX root =Feather.tex

%%\vspace{-2mm}
\subsection{Private Set Intersection (PSI)}\label{sec::psi}

PSI is a cryptographic protocol that enables mutually distrustful parties to compute the intersection of their private datasets without revealing anything about the datasets beyond the intersection. 

The fundamental functionality computed by any $n$-party PSI can be defined as $\mathcal{G}$ which takes as input sets $S_{\st 1},..., S_{\st n}$ each of which belongs to a party and returns the intersection $S_{\st \cap}$ of the sets to a party. More formally, the functionality is defined as:  $\mathcal{G} (S_{\st 1},..., S_{\st n})\rightarrow(S_{\st\cap},\underbrace{\empt,..., \empt}_{\st n-1})$, where $S_{\st\cap}= S_{\st 1} \cap S_{\st 2}, ...,\cap\  S_{\st n}$. In this work, we denote the concrete PSI protocol with $\mathcal{PSI}$.

% PSIs can be categorized into traditional and delegated types. In traditional PSIs, data owners interactively compute the results using their local data, whereas delegated PSIs utilize cloud computing for computation and/or storage while preserving the privacy of the computation inputs and outputs from the cloud. 

%\vspace{-2mm}
\subsection{Local Differential Privacy}\label{sec::local-differential-privacy}

Local Differential Privacy (LDP) entails that the necessary noise addition for achieving differential privacy is executed locally by each individual. Each individual employs a random perturbation algorithm, denoted as $M$, and transmits the outcomes to the central entity. The perturbed results are designed to ensure the protection of individual data in accordance with the specified $\epsilon$ value. This concept has been formally stated in ~\cite{duchi2013local}. Below, we restate it. 

\newtheorem{ldp-definition}{Definition}
\begin{ldp-definition}
   Let X be a set of possible values and Y the set of noisy values. $M$ is $\epsilon$-locally differentially private ($\epsilon$-LDP) if for all $x, x'$ $\in X$ and for all $y \in Y$:
   \begin{equation}
	Pr[M(x)=y]\leq e^{\st\epsilon} \cdot Pr[M(x')=y]
   \end{equation}
 \end{ldp-definition}

For a binary attribute, i.e., $X = \{0, 1\}$, this protection means that an adversary who observes $y$ cannot be sure whether the true value was 0 or 1. 

As proposed by Wang \et. \cite {wang2016using}, we consider two generalized mechanisms on binary attributes for achieving LDP. The first one uses the Randomized Response (RR) and the second one relies on adding Laplace noise with post-processing (applying a threshold of $0.5$) for binarizing the values. For either mechanism, each individual employs a $2\times2$ transformation matrix $P = [p_{\st ij}]$ to perturb their true value, where the element at position $(i,j)$ represents the probability of responding with value $j$ if the true value is $i$. To satisfy the definition of DP at privacy level $\epsilon$, we need to have $p_{\st 00}/p_{\st 01} \leq \epsilon$. 

\subsubsection{Randomized Response.}\label{sec::randomised-response}
In addition to the requirement of satisfying $\epsilon$-LDP,  Wang \et. \cite {wang2016using} propose selecting the matrix parameters to maximize the probability of retaining the true value, i.e., to maximize $p_{\st 00} + p_{\st 11}$. This yields the following transformation matrix:

\begin{equation}\label{eq:rr}
  Q:=  \begin{pmatrix}
\frac{e^{\st\epsilon}}{1+ e^{\st\epsilon}} & \frac{1}{1+ e^{\st \epsilon}} \\
\frac{1}{1+ e^{\st\epsilon}} & \frac{e^{\st\epsilon}}{1+ e^{\st\epsilon}}
\end{pmatrix}
\end{equation}

\subsubsection{Laplace Noise with Post-Processing.} \label{sec::laplace-noise}
The Laplace mechanism is a DP mechanism proposed by the original DP paper~\cite{dwork2006calibrating}. To achieve $\epsilon$-DP, this mechanism adds noise drawn from the Laplace distribution with parameter $\frac{1}{\epsilon}$ to the true value. This creates continuous values, instead of binary ones. Consequently, we need to make the output binary. 
It is demonstrated in \cite {wang2016using} that using a threshold of 0.5 maximizes $p_{\st 00} + p_{\st 11}$, i.e., if the continuous value is above 0.5, we set the final value to 1; otherwise, we set it to 0. This leads to the following transformation matrix:

\begin{equation}\label{eq:laplace}
Q':=\begin{pmatrix}
1- \frac{1}{2}e^{\st-\frac{\epsilon}{2}} & \frac{1}{2}e^{\st-\frac{\epsilon}{2}} \\
\frac{1}{2}e^{\st-\frac{\epsilon}{2}} & 1- \frac{1}{2}e^{\st-\frac{\epsilon}{2}}
\end{pmatrix}
\end{equation}

Note that although we list the matrices only for binary attributes here, both mechanisms generalize to the case of categorical variables with more than two values.

\subsubsection{Mechanisms for Optimal Inference Privacy.}\label{sec::game-matrices}

Any randomization mechanism for obfuscating the flags while sharing can offer certain protection against inference attacks by \aut. Given a value of $\epsilon$, there can be many mechanisms that satisfy the constraint of DP, of which two can be found using Equation~\ref{eq:rr} (for RR) and Equation~\ref{eq:laplace} (for Laplace). 

These mechanisms assign an equal probability of converting a 0 to 1 and 1 to 0. They need not be the optimal transformation matrices that provide maximum inference privacy, i.e., maximum protection against \aut's ability to infer the flag values. 

As one of the key contributions of this work, { we developed a framework to explore the entire space of transformation matrices and find optimal mechanisms that maximize inference privacy, under the given constraints on utility and local differential privacy}.

 The main advantage of formulating the construction of a privacy mechanism as an optimization problem is that we can automatically explore a large solution space to discover optimal mechanisms that are not expressible in closed form (such as the Laplace or Gaussian mechanism).  Section \ref{sec:formulation} presents further details about the game construction and solution.

%\vspace{-2mm}

\subsection{Federated Learning}\label{sec:Federated-Learning}

% Federated Learning (FL) introduced in  \cite{mcmahan2017communication} is an innovative machine learning approach designed to train models across decentralised entities while keeping data localised. 

Unlike traditional centralized methods, where data is pooled into a central server, FL allows model training to occur on individual devices/clients contributing private data. This preserves the privacy of the data to some extent by avoiding direct access to them. The process involves training a global model through collaborative learning on local data, and only the model updates, rather than raw data, are transmitted to the central server. 

This decentralized paradigm is particularly advantageous in scenarios where data privacy is paramount, such as in healthcare or finance, as it enables machine learning advancements without compromising sensitive information.
Algorithm~\ref{flalgorithm} presents the overall workflow of FL.

%\vspace{-2.2mm}
\begin{algorithm}[H]
\caption{: Federated Learning's General Procedure}\label{flalgorithm}
\begin{algorithmic}[1]
\State \textbf{Server:}
\State Initialize global model $\theta$
\For{each round $k = 1,2,3,..., K$}
    \State Broadcast $\theta$ to all participating devices
    \State \textbf{Clients:}
     \For{each client $i$ (where $1\leq i\leq n$) in parallel}
        \State Receive global model $\theta$
        \State Compute local update $g_{\st i}$ using local data
        \State Send $g_{\st i}$ to the server
    \EndFor
    \State \textbf{Server:}
    \State Aggregate local updates: $G_{\st k} = \sum\limits_{\st i=1}^{\st n} g_{\st i}$
    \State Update global model: $\theta_{\st k+1} = \text{UpdateModel}(\theta_{\st k}, G_{\st k})$
\EndFor
\end{algorithmic}
\end{algorithm}

%\vspace{-3.3mm}
\subsubsection{SecureBoost: A Lossless Vertical Federated Learning
Framework.}\label{sec:SecureBoost}

SecureBoost, introduced in \cite{cheng2021secureboost}, stands out as an innovative FL framework designed to facilitate collaborative machine learning model training among multiple parties while safeguarding the privacy of their individual datasets. It accomplishes this by leveraging homomorphic encryption to execute computations on encrypted data, ensuring the confidentiality of sensitive information throughout the training procedure.
There are two main technical concepts and phases involved in SecureBoost:

\

%\vspace{-.5mm}
\begin{itemize}[leftmargin=4mm]
    \item[$\bullet$]{\textbf{Secure Tree Construction:}} SecureBoost builds boosting trees, a specific type of machine learning model, by utilizing a non-federated tree boosting mechanism called XGBoost \cite{ChenG16} and a partially homomorphic encryption scheme, such as Paillier encryption \cite{Paillier99}, allowing various operations such as majority votes and tree splits to be performed without exposing the underlying plaintext data to the system's participants.
    \item[$\bullet$]{\textbf{Entity Alignment:}} To enable collaborative training, SecureBoost conducts entity alignment to recognize corresponding user records across diverse data silos. This process is typically executed through an MPC (such as PSI), guaranteeing the confidentiality of individual identities.
\end{itemize}

SecureBoost has been implemented in an open-sourced FL project, called FATE.\footnote{\url{https://github.com/FederatedAI/FATE}} As discussed above, (V)FL is an interactive process within which parties exchange messages. Thus, there is a possibility of a leakage to these parties. To formally define the leakage to each party in this process, below we introduce a leakage function $\leakW$.
\begin{equation}\label{def::leakW}
    \leakW(\prm_{\st 1},..., \prm_{\st n} )\rightarrow (l_{\st 1},..., l_{\st n})
\end{equation}

This function receives the input parameter $\prm_{\st i}$ from each party in (V)FL and returns  leakage $l_{\st i}$ to the i-th party, representing the information that (V)FL exposes to that specific party. Note that $\prm_{\st i}$ is a set, containing all (intermediate) results possibly generated over multiple iterations. This leakage will be considered in \starlit's formal definition (in Sections \ref{sec:Formal-Security-Definition} and \ref{sec::sec-analysis}) and proof (in Appendix \ref{sec::Proof-of-Security}).

%\vspace{-2mm}
\subsection{\flower: A Federated Learning Implementation Platform}\label{sec:flower}

We implement \starlit within \textit{\flower}, which was introduced in \cite{abs-2007-14390}. This framework offers several advantages, including scalability, ease of use, and language and ML framework agnosticism.

\flower comprises three main components: a set of clients, a server, and a strategy. Federated learning is often viewed as a combination of global and local computations. The server handles global computations and oversees the learning process coordination among the clients. The clients perform local computations, utilizing data for training or evaluating model parameters.

% Thus, the client and the server implement the basic functionalities of the clients and server in federated learning training. 

The logic for client selection, configuration, parameter update aggregation, and federated or centralized model evaluation can be articulated through strategy abstraction.
The implementation of the strategy represents a specific FL algorithm. \flower provides reference implementations of popular FL algorithms such as FedAvg \cite{mcmahan2017communication}, FedOptim \cite{abs-2003-00295}, or FedProx \cite{LiSZSTS20}.

%\vspace{-2mm}
\subsection{Security Model}\label{sec::sec-model}

In this paper, we consider static adversaries. We use the simulation-based paradigm of secure multi-party computation \cite{DBLP:books/cu/Goldreich2004} to define and discuss the security of the proposed scheme. Since we focus on the static passive (semi-honest) adversarial model, we will restate the security definition in this adversarial model. 
%
 %\vspace{-1mm}
 \subsubsection{Two-party Computation.} A two-party protocol $\Gamma$ problem is captured by specifying a random process that maps pairs of inputs to pairs of outputs, one for each party. Such process is referred to as a functionality denoted by  $\func:\{0,1\}^{\st  *}\times\{0,1\}^{\st  *}\rightarrow\{0,1\}^{\st  *}\times\{0,1\}^{\st  *}$, where $\func := (\func_{\st  1},\func_{\st  2})$. For every input pair $(x,y)$, the output pair is a random variable $(\func_{\st  1} (x,y), \func_{\st  2} (x,y))$, such that the party with input $x$ wishes to obtain $\func_{\st  1} (x,y)$ while the party with input $y$ wishes to receive $\func_{\st  2} (x,y)$. When $\func$ is deterministic, then $\func_{\st  1} = \func_{\st  2}$. The above functionality can be easily extended to $n>2$ parties.

 %In the setting where $f$ is asymmetric and only one party (say the first one) receives the result, $f$ is defined as $f:=(f_{  1}(x,y), \emptyset)$. 
 
 %\vspace{-1mm}
 \subsubsection{Security in the Presence of Passive Adversaries.}  In the passive adversarial model, the party corrupted by such an adversary correctly follows the protocol specification. Nonetheless, the adversary obtains the internal state of the corrupted party, including the transcript of all the messages received, and tries to use this to learn information that should remain private. 
 
 Loosely speaking, a protocol is secure if whatever can be computed by a party in the protocol can be computed using its input and output only. In the simulation-based model, it is required that a party’s view in a protocol's 
 execution can be simulated given only its input and output. This implies that the parties learn nothing from the protocol's execution. More formally, party $i$’s view (during the execution of $\Gamma$) on input pair  $(x, y)$ is denoted by $\mathsf{View}_{\st  i}^{\st  \Gamma}(x,y)$ and equals $(w, r^{\st  i}, m_{\st  1}^{\st  i}, ..., m_{\st  t}^{\st  i})$, where $w\in\{x,y\}$ is the input of $i^{\st  th}$ party, $r_{\st  i}$ is the outcome of this party's internal random coin tosses, and $m_{\st  j}^{\st  i}$ represents the $j^{\st  th}$ message this party receives.  The output of the $i^{\st  th}$ party during the execution of $\Gamma$ on $(x, y)$ is denoted by $\mathsf{Output}_{\st  i}^{\st  \Gamma}(x,y)$ and can be generated from its own view of the execution.  %The joint output of both parties is denoted by $\mathsf{Output}^{  \Gamma}(x,y):=(\mathsf{Output}_{  1}^{  \Gamma}(x,y), \mathsf{Output}_{  2}^{  \Gamma}(x,y))$.
%
%\vspace{-2mm}
\begin{definition}\label{def::simulation-based-def-semi-honest}
Let $\func$ be the deterministic functionality defined above. Protocol $\Gamma$ securely computes $\func$ in the presence of a  passive probabilistic polynomial-time (PPT) adversary \adv, if for every \adv in the real model, there exist PPT algorithms $(\mathsf {Sim}_{\st  1}, \mathsf {Sim}_{\st  2})$ such that:
\end{definition}
%
%\vspace{-2mm}
  \begin{equation*}
  \{\mathsf {Sim}_{\st 1}(x, \func_{\st 1}(x,y))\}_{\st x,y}\stackrel{c}{\equiv} \{\mathsf{View}_{\st  1}^{\st \adv, \Gamma}(x,y) \}_{\st x,y}
  \end{equation*}
  \begin{equation*}
    \{\mathsf{Sim}_{\st 2}(y, \func_{\st 2}(x,y))\}_{\st x,y}\stackrel{c}{\equiv} \{\mathsf{View}_{\st 2}^{\st \adv, \Gamma}(x,y) \}_{\st x,y}
  \end{equation*}

Definition \ref{def::simulation-based-def-semi-honest} can be easily extended to $n>2$ parties.

%\newpage % only for ease when writing, to be removed

%\vspace{-2mm}
\section{System Design} \label{sec:solution-design}

% Banks sharing only binary values (about whether the information at account and transaction levels is anomalous or not) complies with the principle of {\bf data minimisation}, as this data is adequate, necessary, and sufficient for achieving the objective at hand. To align with the principle of {\bf integrity and confidentiality}, we compute the transaction-level discrepancy features using a {\it secure private set intersection} protocol and protect the flag values using {\it local differential privacy}. 

% \begin{tcolorbox}
% {\it Our system ensures that the data at the banks (such as names and addresses) is never disclosed in plain text form to \fsp. And by design, banks only share minimal information that is necessary for improving anomaly detection models. Moreover, for a desired level of improvement in accuracy, our framework finds and applies the mechanism that provides the maximum protection to flag values at the banks.}
% \end{tcolorbox}

% \sasi{Need a statement here that satisfies the criterion of "explaining the guarantees to the regulators"..}

%\subsection{Solution Architecture:} \label{sec:solution-overview}
%\sasi{I think this section needs an}

%{\bf Solution Architecture: } 

\starlit consists of two main phases: (i) feature extraction and (ii) training. 
During the feature extraction phase, 
the two types of features (discussed in Section \ref{sec::problem-statement}) are retrieved in a privacy-preserving manner, the data is aligned, and then passed onto a third party, called ``Feature Collector (\fc)''. The use of \fc drastically simplifies the training phase from $n$-party down to $2$-party VFL, which will enable the system to scale to a large number of banks.

\begin{figure}[ht]
%\vspace{-2mm}
    \centering
\includegraphics[width=.47\textwidth]{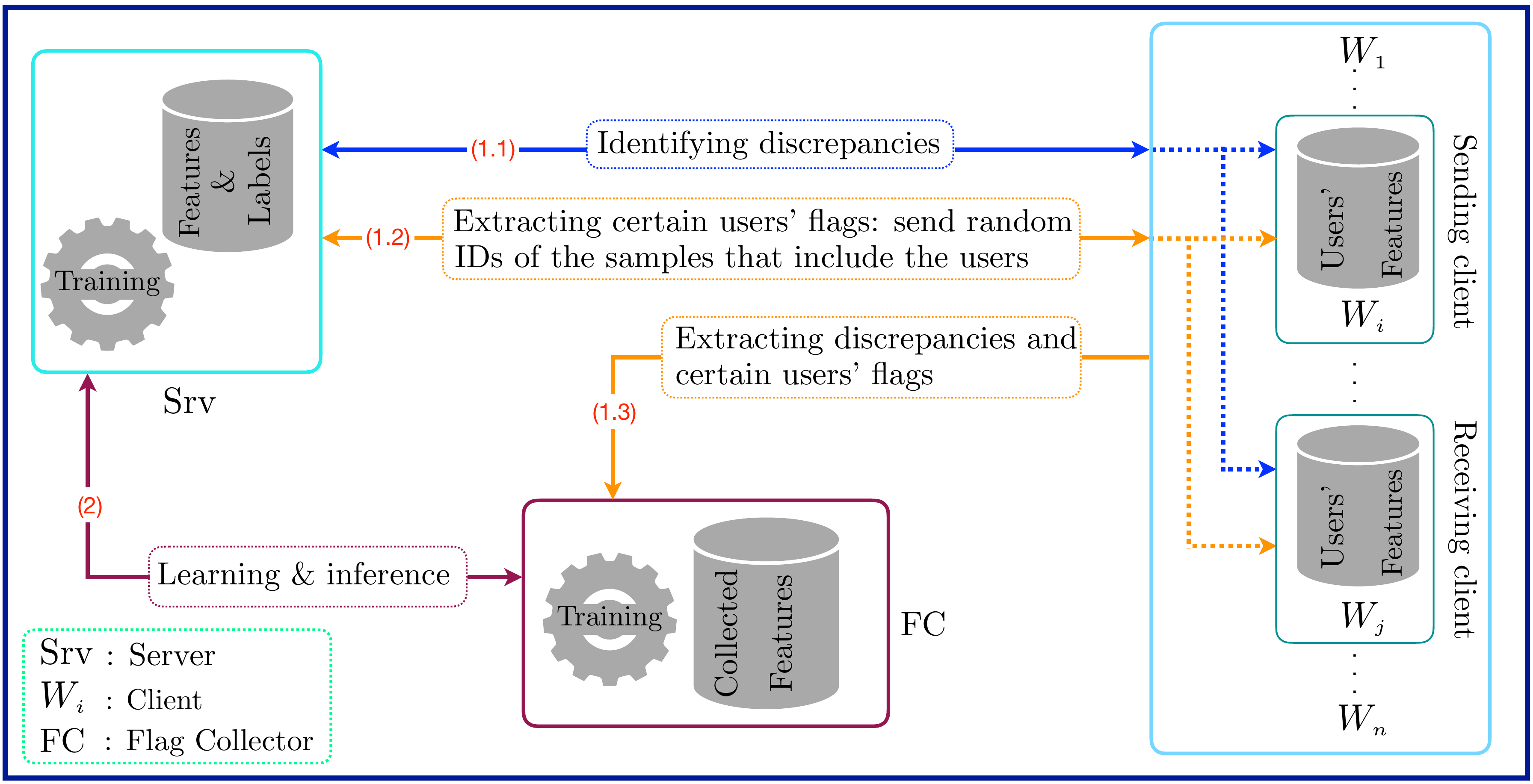}
    \caption{\small Outline of parties'  interactions in \starlit.}
    \label{fig:high-level-architecture}
    %\vspace{-3mm}
\end{figure}

Figure \ref{fig:high-level-architecture} outlines the interaction between the parties in \starlit. In Phase 1, 
each client initially engages with \aut to identify discrepancies in specific user features. Additionally, in the same phase, each client interacts with \aut to extract flags for certain users. Subsequently, each client combines the results of discrepancy extraction with the outcomes of flag extraction, sending the pair along with a random ID (known also to \aut) to \fc.  
Moving on to Phase 2, \fc and \aut collaborate to train the VFL model using \fc's collected features, \aut's local data, and SecureBoost.

%\aut assigns to each user a random ID if needed. In certain cases, such as a financial transaction, each transaction already is accompanied by a random unique ID. \aut transmits to its partner bank this ID and minimal information about the user that its partner bank holds, e.g., the user's account number or name. 
%
%Furthermore, banks only transmit the ID and flags that are paired with these account numbers to \fc. %Neither the banks nor \fc can make useful inferences from these messages. 

This procedure may still leave the chance of an inference attack during model training/deployment. To address this issue, we use LDP, where any flag values that leave the client are obfuscated via a randomization strategy. %, providing the strongest possible privacy guarantees.  
%
% Details on the mechanisms we used to achieve local differential privacy for the flag values are provided below. 
%
Note that this protection is an additional layer on top of what is already offered by the SecureBoost protocol, which only shares encrypted (aggregated) gradient information.

\section{Formal Security Definition}\label{sec:Formal-Security-Definition}

In this section, we introduce a generic formal definition, that we call \cel. It establishes the primary security requirements of privacy-preserving (V)FL schemes such as \starlit. 
\cel involves three types of parties, (i) a service provider \aut, (ii) a feature collector \fc, and (iii) a set of clients $\{\bank_{\st 1},..., \bank_{\st n}\}$ contributing their private inputs. 
Informally, \cel allows \aut to generate a (global) model given its initial model and the inputs of $\bank_{\st 1},..., \bank_{\st n}$. To achieve a high level of computational efficiency and scalability, in \cel, we involve a third-party \fc that assists \aut with computing the model (by interacting with $\bank_{\st i}$s and retrieving the features they hold). 
The functionality $\func$ that \cel computes takes an input initial model $\theta$ from \aut, a set $S_{\st i}$ from every  $\bank_{\st i}$, and no input from \fc. It returns to \aut an updated model $\theta'$. It returns nothing to the rest of the parties.\footnote{For the sake of simplicity, we have restricted the learning of the global model to only \aut. This approach can be easily generalized to allow each $\bank_{\st i}$ to learn the model as well, by mandating \aut to transmit the global model to every $\bank_{\st i}$.}  Hence, $\func$ can be formally defined as follows. 
\begin{equation}\label{equ::func-cel}
\func(\theta, S_{\st 1},..., S_{\st n}, \empt)\rightarrow (\theta',\underbrace{\empt,...,\empt}_{\st n},\empt)
\end{equation}

%\vspace{-1mm}
Since (i) \fc interacts with $\bank_{\st 1},..., \bank_{\st n}$ and collects some features from them and (ii) \aut generates the model in collaboration with $\bank_{\st 1},..., \bank_{\st n}$ and \fc,  there is a possibility of leakage to the participating parties. Depending on the protocol that realizes $\func $ this leakage could contain different types of information. For instance, it could contain (a) each $\bank_{\st i}$'s local model outputs and corresponding gradients (a.k.a. intermediate results) when using gradient descent \cite{WanNHL07} in VFL, (b) the output of entity aligning procedure, (c) information about features, or (d) nothing at all. We define this leakage as an output of a leakage function defined as follows: 
\begin{equation}\label{equ:generic-leak-func}
\leak(inp)\rightarrow(l_{\st 1}, l_{\st 2}, ..., l_{\st  n+2})
\end{equation}

$\leak(inp)$ takes all parties (encoded) inputs, denoted as $inp$. It returns leakage $l_{\st 1}$ to \aut, $l_{\st 2}$ to \fc, and leakage $l_{i}$ to $\bank_{\st i-2}$, for all $i$, where $3\leq i\leq n+2$.

%
% To clarify, $\leak$'s output  $l_{\st P}$ is given to party $P$. 
%
% In this paper, by  $\leak_{\st P}(inp)$ we refer to the output of $\leak(inp)$ that is returned to party $P$, i.e.,  $l_{\st P}$.  

We assert that a protocol securely realizes $\func$ if (1) it reveals nothing beyond a predefined leakage to a certain party and (2) whatever can be computed by a party in the protocol can be obtained from its input and output only. This is formalized by the simulation paradigm. We require a party's view during the execution of the protocol to be simulatable given its input, output, and the leakage that has been defined for that party.

 \begin{definition}[Security of \cel]\label{def::cel-def} Let $\func$ be the functionality presented in Relation \ref{equ::func-cel}. Also, let $\leak$ be the above leakage function, presented in Relation \ref{equ:generic-leak-func}. We assert that protocol $\Gamma$ securely realizes $\func$, in the presence of a static semi-honest adversary, if for every non-uniform PPT adversary \adv for the real model, there exists a non-uniform PPT adversary (or simulator) 
 
 % $\mathsf{Sim}^{\st \func, \leak_{\st P}}_{\st P}$
 %
 $\mathsf{Sim}$ for the ideal model, such that for every party $P$, where $P\in\{\aut, \bank_{\st 1},..., \bank_{\st n}, \fc\}$, the following holds: 
 
 %\vspace{-2mm}
  \begin{equation}\label{equ::client-sec}
  \{\mathsf{Sim}^{\st \func,  \leak_{\st 1}}_{_{\st \aut}}(\theta, \theta')\}_{\st inp}\stackrel{c}{\equiv} \{\mathsf{View}_{\st \aut}^{\st \adv, \Gamma}(inp) \}_{\st inp}
  \end{equation}
    \begin{equation}\label{equ::tee-sec}
    \{\mathsf{Sim}_{\st\fc}^{\st\func, \leak_{\st 2}}(\empt, \empt)\}_{\st inp}\stackrel{c}{\equiv} \{\mathsf{View}_{\st \fc}^{\st \adv, \Gamma}(inp) \}_{\st inp}
  \end{equation}
  \begin{equation}\label{equ::tee-sec}
    \{\mathsf{Sim}_{\st\bank_{i}}^{\st \func, \leak_{\st i+2}}(S_{\st i}, \empt)\}_{\st inp}\stackrel{c}{\equiv} \{\mathsf{View}_{\st \bank_{i}}^{\st \adv, \Gamma}(inp) \}_{\st inp}
  \end{equation}
%

%   \begin{equation}\label{equ::client-sec}
%   \{\mathsf{Sim}^{\st \func,  \leak_{\st \fsp}}_{_{\st \fsp}}(\theta, \theta')\}_{\st inp}\stackrel{c}{\equiv} \{\mathsf{View}_{\st \fsp}^{\st \adv, \Gamma}(inp) \}_{\st inp}
%   \end{equation}
%   %
%   \begin{equation}\label{equ::tee-sec}
%     \{\mathsf{Sim}_{\st\bank_{i}}^{\st \func, \leak_{\st \bank_{i}}}(S_{\st i}, \empt)\}_{\st inp}\stackrel{c}{\equiv} \{\mathsf{View}_{\st \bank_{i}}^{\st \adv, \Gamma}(inp) \}_{\st inp}
%   \end{equation}
% %
%     \begin{equation}\label{equ::tee-sec}
%     \{\mathsf{Sim}_{\st\fc}^{\st\func, \leak_{\st \fc}}(\empt, \empt)\}_{\st inp}\stackrel{c}{\equiv} \{\mathsf{View}_{\st \fc}^{\st \adv, \Gamma}(inp) \}_{\st inp}
%   \end{equation}

where $1\leq i \leq n$.
\end{definition}

%\vspace{-3mm}
\section{Flag Protection}\label{sec:formulation}
%\vspace{-1mm}
In this section, we initially present a game for flag protection. Then, we explain how to construct a concrete optimization problem to realize the game. 

 %%\vspace{-2mm}
\subsection{The Game} 
The problem of finding a Privacy Mechanism (PM) that offers optimal flag privacy to a client given the knowledge of the adversary (e.g., \aut or \fc), is an instance of a Bayesian Stackelberg game. In a Stackelberg game the \textit{leader}, in our case the client, plays first by choosing a PM (a transformation matrix), and commits to that by running it on the actual values of the flags; and the \textit{follower}, in our case \aut, plays next estimating the flag value, knowing the PM that the client has committed to. It is a Bayesian game because \aut has incomplete information about the true flag values and plays according to its prior information about these values. Inspired by similar work in location privacy protection games~\cite{shokri2012protecting,shokri2014privacy}, we now proceed to define the game for a single flag value, but the transformation matrix computed will be used for each value:

%%%%%%%%%%%%%%%%%%%%%%%%%%%%%%%%%%%%

\begin{figure}[ht]%[!htbp]
%\vspace{-4mm}
\setlength{\fboxsep}{1pt}
\begin{center}
    \begin{tcolorbox}[enhanced, 
    drop fuzzy shadow southwest,
    colframe=black,colback=white,height=6.2cm]
    %\begin{tcolorbox}[enhanced]%, colback=white, drop shadow={black,opacity=1}]
    % {\small{
%%********************
%\vspace{-2mm}
\begin{itemize}[leftmargin=-1mm,label=$\bullet$]
    \item \textbf{Step 0.} Nature selects a flag value $v \in \set{V}$ for the client according to a probability distribution $\pi(.)$, the \textit{flag profile}. That is, flag value $v$ is selected with probability $\pi(v)$. This encodes the relative proportions of the flag values in the dataset.
    \item \textbf{Step 1.} Given $v$, the client runs the PM $f(v'|v)$ to select a replacement value $v' \in \set{V}$.
    \item \textbf{Step 2.} Having observed $v'$, \aut selects an estimated flag value $\hat{v}\sim g(\hat{v}|v'), \hat{v} \in \set{V}$. \aut knows the probability distribution $f(v'|v)$ used by the PM, and the client's flag profile $\pi(.)$, but not the true flag value $v$.
    \item \textbf{Step 3.} The game outcome is the number $\deltap(\hat{v},v)$, which is the client's privacy for this iteration of the game. This number represents \aut's error in estimating the true value of the flag.
\end{itemize}
%%*******
%}
%}
\end{tcolorbox}
%\end{cryptoProtocolBox}
\end{center}
%\vspace{-1mm}
\caption{Bayesian game for a single flag value.}
\label{fig:DP-game}
 %\vspace{-3mm}
\end{figure}
%%%%%%%%%%%%%%%%%%%%%%%%%%%%%%

%\vspace{-1mm}

%%%%%*************%%%%
The above description is common knowledge to \aut and the client. \aut tries to minimize the expected game outcome (the error in the estimation of the flag value) via its choice of $g$, while the client tries to maximize it via its choice of transformation matrix $f$. %, subject to the expected quality loss being less than a given threshold $\dqmax$.
As changing the flag values distorts the data for training the ML algorithm, we impose upper bounds $p^{\max}(v',v)$ on the probabilities $f(v'|v)$. 
Finally, and independently of the above considerations, we want the PM to be $\epsilon$-differentially private.

%\vspace{-1mm}
\subsection{Optimization Problem}\label{sec:solution}
%%\vspace{-1mm}
We now explain how to build a concrete optimization problem that encodes the above description and that we can solve to obtain the optimal PM $f()$, given $\pi(), \deltap, p^{\st\max}(v',v)$, and $\epsilon$. 
%, \deltaq, \dqmax$.
%
\aut knows $f(v'|v)$ implemented by PM. Thus, it can form a posterior distribution $\Pr(v|v')$ on the true flag value, conditional on the observation $v'$. Then, \aut chooses $\hat{v}$ to minimize the conditional expected privacy, where the expectation is taken under the posterior distribution:
%
%\vspace{-1mm}
\begin{equation}\label{eq:advobjective}
   \text{Choose $\hat{v}$ that satisfies } \arg \min_{\st\hat{v}}\sum_{\st v} \Pr(v|v') \deltap(\hat{v}, v).
\end{equation}

Recall that variables $v, v',$ and $ \hat{v}$  take values in $\set{V}$, the set of flag values, so the range of any minimization or summation involving any of these variables will be the set $\set{V}$. 
If there are multiple minimizing values of $\hat{v}$, then \aut may randomize among them. This randomization is expressed through $g(\hat{v}|v')$, and in this case \eqref{eq:advobjective} would be rewritten as 
%\begin{equation}\label{eq:advobjective-alt}
$\sum_{\st v,\hat{v}} \Pr(v|v')g(\hat{v}|v') \deltap(\hat{v}, v)$.
    
%\end{equation}

It is important to note that the value of this equation would be the same as the value computed in Relation \eqref{eq:advobjective} for any minimizing value of $\hat{v}$. 
As $\pi(v)$ and $f(v|v')$ are known to \aut, it holds that:  
\begin{equation}
    \Pr(v|v') = \frac{\Pr(v, v')}{\Pr(v')}=\frac{f(v'|v)\pi(v)}{\sum_{\st v} f(v'|v)\pi(v)}
\end{equation}

Thus, for a given $v'$, the client's conditional privacy is given by Relation \eqref{eq:advobjective}. The probability that $v'$ is reported is $\Pr(v')$. Hence, the unconditional expected privacy of the client is: 
\begin{equation}\label{eq:userEprivacy-user}
    \sum_{\st v'}\Pr(v')\min_{\st \hat{v}}\sum_{\st v} \Pr(v|v') \deltap(\hat{v}, v)
    = \sum_{\st v'}\min_{\st \hat{v}}\sum_{\st v} \pi(v)f(v'|v) \deltap(\hat{v}, v)
\end{equation}

To facilitate computations, we define:
\begin{equation}\label{def:xr}
    x_{\st v'} \triangleq \min_{\st \hat{v}}\sum_{\st v} \pi(v)f(v'|v) \deltap(\hat{v}, v).
\end{equation}
Incorporating $x_{\st v'}$ into Relation \eqref{eq:userEprivacy-user}, the unconditional expected privacy of the client can be rewritten as
\begin{equation}\label{eq:userEprivacy-x}
    \sum_{\st v'}x_{\st v'}
\end{equation}
which the client aims to maximize by choosing $f(v'|v)$. The minimum operator makes the problem non-linear, undesirable, but 
 Relation \eqref{def:xr} can be transformed into a series of linear constraints:
\begin{equation}\label{eq:xr-linear}
    x_{\st v'} \leq \sum_{\st v} \pi(v)f(v'|v) \deltap(\hat{v}, v), \forall \hat{v}
\end{equation}

Maximizing the result in Relation \eqref{eq:userEprivacy-x} under 
 Relation \eqref{def:xr} is equivalent to maximizing Relation \eqref{eq:userEprivacy-x} under Relation \eqref{eq:xr-linear}. For every $v'$, there must be some $\hat{v}$ for which Relation \eqref{eq:xr-linear} holds as strict equality; Otherwise, we could increase one of the $x_{\st v'}$, so the value of Relation \eqref{eq:userEprivacy-x} would increase. 
From Relations \eqref{eq:userEprivacy-x} and \eqref{eq:xr-linear}, the linear program for the client is constructed by choosing $f(v' | v), x_{\st v'}, \forall v, v'$ to solve the following linear programming problem.
%\vspace{-.5mm}
\begin{align}
    \text{\textbf{Maximize}} & \sum_{\st v'} x_{\st v'} \label{obj_u}\\
    \text{\textbf{subject to}}& \nonumber\\
    & x_{\st v'} - \sum_{\st v} \pi(v) f(v' | v) \deltap(\hat{v}, v) \leq 0, \forall \hat{v}, v' \label{sub_u1}\\
    %& \sum_v \pi(v)\sum_{v'} f(v'|v) \deltaq(v', v) \leq \dqmax \label{sub_u2}\\
    & f(v' | v) \leq p^{\st\st\max}(v',v), \forall v, v' \label{sub_u2}\\
    & \sum_{\st v'} f(v' | v) = 1, \forall v \label{sub_u3}\\
    & f(v' | v) \geq 0, \forall v, v' \label{sub_u4}\\
    & \frac{f(v' | v_{\st 1})}{f(v' | v_2)} \leq \exp(\epsilon), \forall v', v_1, v_2 \label{sub_u5}
\end{align}

%Inequality \eqref{sub_u2} reflects the quality loss constraint;
Constraints \eqref{sub_u3} and \eqref{sub_u4} reflect that $f(v' | v)$ is a probability distribution function for each $v$, while  \eqref{sub_u5} enforces $\epsilon$-differential privacy.

%\vspace{-1mm}
\subsubsection{Alternative Quality-Privacy Tradeoffs.} The above formulation encodes the privacy-accuracy tradeoff in one particular way -- maximize inference privacy, subject to a differential privacy constraint and an accuracy-related constraint on the probabilities $f (v'|v)$. The general framework is flexible to accommodate other tradeoffs.

For example, instead of introducing constraints $p^{\st\max}(v',v)$ on $f(v' | v)$, we can introduce an Accuracy Loss (AL) matrix with entries $AL_{\st v'v}$ that quantify the loss in accuracy when replacing value $v$ with $v'$. Then, instead of Relation \eqref{sub_u2}, we can upper bound the total expected accuracy loss that is caused by a given transformation matrix $f$ with the following constraint:
%\vspace{-1mm}
\begin{displaymath}
AL(f) := \sum_{\st v} \pi(v)\sum_{\st v'} f(v'|v) AL_{\st v'v} \leq AL^{\st\max}.
\end{displaymath}

Alternatively, in a more radical departure from the original formulation, rather than aiming to maximize the client's privacy (inference privacy) subject to AL constraints, we could instead aim to minimize the accuracy loss $AL(f)$ subject to a lower bound on inference privacy, i.e., $\sum_{\st v'} x_{\st v'} \geq PR^{\st\min}$.

In general, the main benefit of formulating the construction of the transformation matrix as an optimization problem is that we can automatically explore a large solution space to discover optimal probability distributions $f(v'|v)$ that are not expressible in closed form (such as the Laplace or Gaussian mechanism), so human intuition would not be able to find them.

% \section{Randomized response}

% We consider the modern version of Randomized response: Flip a coin -- if the outcome is Tails, respond \texttt{y}; if the outcome is Heads, tell the truth. Assume $\pi$ is the population proportion for whom the true response is 'Y'.

% The attacker knows $\pi$, observes the response $R$ of a user, and aims to infer the true value $V$ of the user. Applying Bayes' law, the attacker obtains the following posterior probabilities for $V$:

% \begin{align}
% 	\Pr(V = \texttt{y} \mid R = \texttt{y}) = & \frac{2\pi}{1 + \pi}\\
% 	\Pr(V = \texttt{n} \mid R = \texttt{y}) = & \frac{1 - \pi}{1 + \pi}\\
% 	\Pr(V = \texttt{y} \mid R = \texttt{n}) = & 0\\
% 	\Pr(V = \texttt{n} \mid R = \texttt{n}) = & 1
% \end{align}

% We compute privacy as the attacker's estimation error, where the error is quantified with $dp()$ and the expectation is taken over the population and over the attacker's random estimation:
% \begin{equation}
% 	\frac{2\pi}{1 + \pi} dp(\hat{y}, y) \pi + \frac{1 - \pi}{1 + \pi} dp(\hat{n}, y)\pi + 0 + 1 dp(\hat{n},n)(1-\pi),
% \end{equation}
% and if we assume that there is zero privacy gain when the attacker makes a correct estimate (i.e. $dp(\hat{y}, y) = dp(\hat{n},n) = 0$), the above formula becomes 
% \begin{equation}
% 	 \pi\frac{1 - \pi}{1 + \pi} dp(\hat{n}, y)
% \end{equation}

% !TEX root =Feather.tex

 %\vspace{-2mm}
\section{Starlit's Phases in Detail}

\subsection{Privacy-Preserving Feature Extraction}\label{sec::feature-extraction}
In this section, we elaborate on the two primary privacy-preserving mechanisms that we designed to extract features. 

%\vspace{-2mm}
\subsubsection{Finding Features' Discrepancies.}\label{sec::comparing-names-and-addresses} 
%\susr
Let $\feas=\{\fea_{\st \susr, 1},..., \fea_{\st \susr, m}\}$ be a subset of features that \aut holds for a  user $\usr$. Consider the scenario where 
%
% \aut wants to check with a pair of clients $(\bank_{\st i}, \bank_{\st j})$ whether the clients have different views of these features. 
%
\aut wants to check with a pair of clients $(\bank_{\st i}, \bank_{\st j})$ if there is a discrepancy between some of the features in $\feas$  that \aut, $\bank_{\st i}$, and $\bank_{\st j}$ hold, without revealing and being able to learn anything else. This approach could provide information about anomalous transactions. 

In the domain of financial transactions, we analyzed synthetic data provided to us and identified key features possessed by \fsp for each transaction (with \fsp acting as \aut). These features include: (i) $customer_{\st name}$, (ii) $countryCity_{\st zipcode}$, and (iii) $street_{\st name}$ for both the ordering and beneficiary banks. Each bank, per user, maintains various features such as $customer_{
\st name}$, $countryCity_{\st zipcode}$, and $street_{\st name}$ (with an associated flag). 

Diverse parties may hold varying perspectives on the value of these features. Discrepancies can arise from various factors. For instance, a user may have supplied divergent information to different parties. In the given scenario, a customer might hold accounts with both the ordering and beneficiary banks but could have provided inconsistent details, such as their address, to these banks. Additionally, there is a possibility that the values maintained by \aut have been tampered with, potentially by external entities  \cite{bergin2016special,Davey-Winder}. Thus, incorporating a feature that signals disparities between a client's data and \aut's data can enhance the accuracy of models. 

To detect discrepancies while preserving privacy we use PSI, a method that safeguards the privacy of non-suspicious users' data maintained by the involved parties. The PSI outcomes serve as additional features in the FL model. Specifically, \aut and each client $\bank_{\st i}$ participate in an instance of PSI, receiving a set of strings from \aut and the client. The PSI 
returns the intersection to $\bank_{\st i}$. For each user, $\bank_{\st i}$ adds a binary feature $b$ to its dataset (if not already present). If a user's details are in the intersection,  $b$ is set to 1; otherwise, it is set to 0. Figure \ref{fig:PSI-used} presents this procedure in detail. Hence, we not only employ PSI (as a subroutine in SecureBoost) for entity alignment, but we also leverage it to enhance the accuracy of the final model. Note that the outcome of the protocol in Figure \ref{fig:PSI-used} will be transmitted to \fc in the second phase (collecting flags of suspicious users), presented below.

%%%%%%%%%%%%%%%%%%%%%%%%%%%%%%%%%%%%
%%\vspace{-1mm}
\begin{figure}[ht]%[!htbp]
%\vspace{-3mm}
\setlength{\fboxsep}{1pt}
\begin{center}
    \begin{tcolorbox}[enhanced, 
    drop fuzzy shadow southwest,
    colframe=black,colback=white,height=7.8cm]
    % \begin{tcolorbox}[enhanced, 
    % drop fuzzy shadow southwest,
    % colframe=black,colback=white]
    % % {\small{
     %\begin{tcolorbox}[enhanced, sharp corners, left=0mm, colback=white, colframe=black, drop shadow={black,opacity=1}, before={%\vspace{-\baselineskip}}]
    %\fbox{\parbox{\linewidth}{%
   % \begin{boxK}
%%********************
%\vspace{-2mm}
\begin{itemize}[leftmargin=-1.3mm,label=$\bullet$]
    \item \textbf{Parties:} \aut and $\bank_{\st i}$. 
    \item \textbf{Input:} 
    \begin{itemize}[leftmargin=1.6mm,label=$\diamond$]
        \item \aut's input, for each user $\usr$, is a set $\feas_{\st \aut}$ of strings (taken from a dataset $DS_{\st \aut}$), where each string has the form $\fea_{\st \susr,1}||\fea_{\st \susr,2}||...||\fea_{\st \susr,m}$ and $\fea_{\st \susr,1}$ is a user's unique ID. 
        
        % $account_{\st number} || customer_{\st name} || street_{\st name} || countryCity_{\st zipcode}$.
        
        \item $\bank_{\st i}$'s input,  for each user $\usr$, is a set $\feas_{\st \bank_{\st i}}$ of strings (from its dataset $DS_{\st\bank_{\st i}}$ of all users), where each string has the form $\fea_{\st \susr,1}||\fea_{\st \susr, 2}||...||\fea_{\st \susr, m}$.
        
        % $account_{\st number} || customer_{\st name} || street_{\st name} || countryCity_{\st zipcode}$.
        
    \end{itemize}
    \item \textbf{Output:} Updated dataset $DS_{\st\bank_{\st i}}$. 
\end{itemize}
 \noindent\rule{\textwidth}{0.1pt}
\begin{enumerate}[leftmargin=1mm]
    \item  \aut and $\bank_{\st i}$ invoke an stance of PSI protocol: $\mathcal{PSI}(\feas_{\st \aut}, \feas_{\st\bank_{\st i}})\rightarrow \feas_{\st \cap}$. 
    \item Given $\feas_{\st \cap}$, $\bank_{\st i}$ parses each element of $\feas_{\st \cap}$ as  
    $\fea_{\st \susr,1}||\fea_{\st \susr,2}||...||\fea_{\st \susr,m}$.
    
    % $(account_{\st number},$ $customer_{\st name},$ $street_{\st name}, countryCity_{\st zipcode})$.
    
    \item  If binary feature $b$ is not in $DS_{\st\bank_{\st i}}$, then $\bank_{\st i}$ adds $b$ to each user's feature. 
    \item $\bank_{\st i}$ sets $b$ as follows. For every $\fea_{\st \susr, j}\in DS_{\st\bank_{\st i}}$:  
    \begin{itemize}[label=$\bullet$]
        \item Sets $b = 1$, when $\fea_{\st \susr, j}\in S_{\st \cap}$. 
        \item  Sets $b = 0$, otherwise. 
    \end{itemize}
    \item $\bank_{\st i}$ returns  $DS_{\st\bank_{\st i}}$. 
\end{enumerate}
\end{tcolorbox}
%}}
%\end{boxK}
\end{center}
%\vspace{-1mm}
\caption{PSI-based method to identify discrepancies.}
\label{fig:PSI-used}
%\vspace{-2mm}
\end{figure}

\subsubsection{Collecting Flags of  Users.}\label{sec::flag-collector} 

Each user's sample may be accompanied by a flag whose value is independently computed and allocated by a client. For instance, in the context of financial transactions, for each user's account that a bank holds, there is a flag indicating whether the bank considers the account suspicious. This flag type offers extra information crucial for anomaly detection. Nevertheless, these flags are treated as private information and cannot be directly shared with \aut. 

To align the flags with the $\aut$'s dataset without revealing them, we rely on the following observation and idea. 
The key observation is that each user's sample, which is held by \aut and includes both sender and receiver clients, can be assigned an ID selected uniformly at random from a sufficiently large domain. In certain cases, such as financial transactions, each sample (representing a transaction) already comes with a random ID. As a random string, this ID divulges no specific information about a user’s features. For each user's sample, \aut can generate this ID and share this ID (along with a unique feature in the sample) with the clients involved in that sample. Accordingly, if each client groups each ID with a set of binary flags and sends them to \fc, \fc cannot glean significant information about the user's features linked to those IDs. Based on this observation, we rely on the following idea to extract the flags.

% The key observation is that each transaction, encompassing both ordering and beneficiary accounts, observed by \fsp, is associated with a unique ID. This ID, being a random string, divulges no specific information about a customer’s account. Therefore, if each bank associates each ID with a (couple of) binary flag(s) and sends these pairs to \fc, \fc cannot glean significant information about the customers' accounts linked to those IDs. Based on this observation, we rely on the following idea to extract the flags.

For each user's sample, \aut sends the random ID and a unique feature of the user (e.g., their name or account number) to the related clients. 
%
% Note that \aut already knows which clients are involved in a sample; therefore \aut only sends to each client limited information about users about whom each client has information. 
%
The clients then use their sample information to group each ID with the correct user's flags. It sends this group to \fc. 
When sending a flag for a user to \fc, each client also sends to \fc the flag $b$ that it generated in Figure \ref{fig:PSI-used} (to detect discrepancies). Consequently, \fc uses a set (that includes an ID and flags for each user) to create a dataset of flags. This dataset will then be used as the input data for the ML model.

% Note that the messages that each bank sends to \fc are encrypted using a private key that they have mutually established beforehand. 

The above private information retrieval mechanism is \textit{highly computationally efficient}. This approach still may reveal certain information to the involved parties. Specifically (a) each client gains knowledge of some of their users that are in \aut's dataset, and (b) \fc acquires information about which IDs originate from certain clients, enabling the calculation of the number of transactions between each pair of clients. 

However, the privacy of sensitive information is preserved, as (i) each client remains unaware of details about other participating clients or users' features held at other clients and (ii) \fc cannot identify the user involved in a sample. \fc only has IDs and a set of flags for each ID. Consequently, \fc cannot glean any information about a specific account.

As evident during the feature extraction, each client independently computes its message and sends it to \fc without the need to coordinate with other clients. Hence, even if some clients choose not to send their messages, this phase is completed. This is in contrast to the solution proposed in \cite{abs-2305-11236} which cannot withstand clients' dropouts.

\noindent\textbf{Extension.} There is an alternative method for collecting flags, which involves employing an efficient \textit{threshold privacy-preserving voting} mechanism introduced by Abadi and Murdoch \cite{AbadiM23}. This voting scheme enables the result recipient  (e.g., \fc or \aut) to ascertain whether, at the very least, a predefined threshold of the involved parties (e.g.,  clients) sets a user's flag to $1$. Importantly, this process does not disclose any additional information, such as individual votes or the count of $1$s or $0$s, beyond the result to the result recipient. This scheme operates with high efficiency, as it avoids the need for public key cryptography. Integrating this scheme in \starlit has the potential to enhance the accuracy of the global model, as there is no longer a requirement to safeguard the flags with DP. A more in-depth analysis is needed to ensure that the system using this voting scheme can withstand potential client dropouts.

For the sake of simplicity, we have presented a solution focused on a single flag per sample. This solution can be readily generalized to situations where multiple flags are linked to a single sample. In this scenario involving multiple flags, when a client receives a sample's ID and the unique user's feature, it retrieves a vector of flags associated with that particular sample. Following applying either DP or the voting-based mechanism to the flags, the client then transmits the resultant outcome to  \fc.

\subsection{Model Training and Inference} \label{sec:: model-learning}

Following the feature extraction phase, 
\aut and \fc jointly possess all the necessary data for training the anomaly detection model. 
\aut retains a dataset of samples, while \fc possesses certain features of samples, i.e., discrepancies and flags (protected by DP). 

This represents the VFL setting, where only \aut holds the labels to predict. This configuration allows for the utilization of various off-the-shelf protocols suitable for training an ML model, such as those presented in \cite{ceballos2020splitnn,cheng2021secureboost,fang2021large,hardy2017private,liu2020asymmetrical,romanini2021pyvertical,sun2021vertical,tian2020federboost,wu2020privacy,xu2021fedv,zhang2020batchcrypt}.  
We use the SecureBoost algorithm (discussed in Section \ref{sec:Federated-Learning}), which involves the exchange of encrypted (aggregate) gradients between \aut and \fc during the training phase. 
\aut can decrypt the gradients to determine the best feature to split on. Once the model is trained, each party owns the part of the tree that uses the features it holds. Hence, when using the distributed inference protocol in \cite{cheng2021secureboost}, \aut coordinates with the \fc to determine the split condition to be used.

\section{Security of  Starlit}\label{sec::sec-analysis}

In this section, we initially present formal definitions of the leakage that each party attains during the execution of \starlit. Subsequently, we formally state the security guarantee of \starlit.

%%%%%%%%%%%%%%%%%%%%%%%%%
\begin{definition}[\aut--Side Leakage]
\label{def:fsp-Side-Leakage} Let $\leak$ be the leakage function defined in Relation \ref{equ:generic-leak-func} and $inp$ be the input of all parties (as outlined in Section \ref{sec:Formal-Security-Definition}). Let $DS_{\st \bank_{\st i}}$ be a dataset of  users held by each $\bank_{\st i}$,  and $v_{\st i}$ be each dataset's size, i.e., $v_{\st i}=|DS_{\st \bank_{\st i}}|$,  where $1\leq i \leq n$. Moreover, let $\leakW(\prm_{\st 1}, \prm_{\st 2})\rightarrow (l_{\st 1}, l_{\st 2})$ be SecureBoost's leakage function (defined in Relation \ref{def::leakW}), where $\prm_{\st 1}$ is provided by \aut and $ \prm_{\st 2}$ is given by \fc. $\leakW$ returns $l_{\st 1}$ to \aut and $l_{\st 2}$ to \fc. Then, leakage to $\aut$ is defined as: 
$\leak_{\st 1}(inp):=\Big(v_{\st 1}, ..., v_{\st n},\leakW_{\st 1}(\prm_{\st 1}, \prm_{\st 2})\Big)$ . 
\end{definition}
%%%%%%%%%%%%%%%%%%%%%%%%%%

%%%%%%%%%%%%%%%%%%%%%%%%%%
\begin{definition}[\fc--Side Leakage]
\label{def:fc-Side-Leakage} Let $\leak$ be the leakage function defined in Relation \ref{equ:generic-leak-func} and $inp$ be the input of all parties.   
 Also, let $s_{\st i}=|S_{\st \bank_{\st i}}|$, where $S_{\st \bank_{\st i}}$ is a set of triples each of which has the form $(ID, b, w)$, where $ID$ represents a random ID of a sample, $b$ is a binary flag for a feature's inconsistency (as described in Figure \ref{fig:PSI-used}), $w$ is another binary flag of the same sample  (as described in Section \ref{sec::flag-collector}). 
Moreover, let $\leakW(\prm_{\st 1}, \prm_{\st 2})\rightarrow (l_{\st 1}, l_{\st 2})$ be SecureBoost's leakage function (defined in Relation \ref{def::leakW}), where $\prm_{\st 1}$ is provided by \aut and $ \prm_{\st 2}$ is given by \fc. $\leakW$ returns $l_{\st 1}$ to \aut and $l_{\st 2}$ to \fc. Then, leakage to $\fc$ is defined as: 
$\leak_{\st 2}(inp):=\Big(s_{\st 1},..., s_{\st n}, \leakW_{\st 2}(\prm_{\st 1}, \prm_{\st 2})\Big)$. 
\end{definition}
%%%%%%%%%%%%%%%%%%%%%%%%%

% %%%%%%%%%%%%%%%%%%%%%%%%%%
% \begin{definition}[\fc--Side Leakage]
% \label{def:fc-Side-Leakage} Let $\leak$ be the leakage function defined in Relation \ref{equ:generic-leak-func} and $inp$ be the input of all parties.   Also, let $S_{\st \bank_{\st i}}$ be a set of triples each of which has the form $(ID, b, w)$, where $ID$ represents a unique transaction random number, $b$ is a binary flag indicating features' inconsistency (computed within the protocol in Figure \ref{fig:PSI-used}), $w$ is a binary flag indicating a suspicious account (computed in the procedure presented in Section \ref{sec::flag-collector}). Moreover, let $\leakW(\prm_{\st 1}, \prm_{\st 2})\rightarrow (l_{\st 1}, l_{\st 2})$ be SecureBoost's leakage function (defined in Relation \ref{def::leakW}), where $\prm_{\st 1}$ is provided by \fsp and $ \prm_{\st 2}$ is given by \fc. $\leakW$ returns $l_{\st 1}$ to \fsp and $l_{\st 2}$ to \fc. Then, leakage to $\fc$ is defined as: 
% %
% $$\leak_{\st 2}(inp):=\Big(\bigcup\limits^{\st i=1}_{\st n}S_{\st \bank_{\st i}}\cup \leakW_{\st 2}(\prm_{\st 1}, \prm_{\st 2})\Big)$$ 
% \end{definition}
% %%%%%%%%%%%%%%%%%%%%%%%%%

%%%%%%%%%%%%%%%%%%%%%%%%
\begin{definition}[$\bank_{\st i}$--Side Leakage] 
\label{def:Bank-Side-Leakage}
Let $\leak$ be the leakage function defined in Relation \ref{equ:generic-leak-func} and $inp$ be the input of all parties. Moreover, let $DS_{\st \aut}$ be \aut's dataset while $DS_{\st \bank_{\st i}}$ be $\bank_{\st i}$'s dataset. Also, let $S_{\st \aut}$ be a set of pairs each of which has the form $(ID, \feat_{\st \susr})$, where $ID$ represents a  random ID of a sample and $\feat_{\st \susr}$ refers to user $\usr$'s unique feature, held by both \aut and $\bank_{\st i}$). Then, leakage to $\bank_{\st i}$  is defined as: 
$\leak_{\st \st i+2}(inp):=\Big((DS_{\st \aut}\ \cap\ DS_{\st \bank_{\st i}}), |DS_{\st \aut}|, S_{\st \aut} \Big)$. 
\end{definition}
%%%%%%%%%%%%%%%%%%%%%%%%%%

\begin{theorem}\label{theo::starlit-privacy}
Let $\func$ be the functionality defined in Relation \ref{equ::func-cel}. Moreover, let
$\leak_{\st 1}(inp), \leak_{\st 2}(inp)$,
and $\leak_{\st \st i+2}(inp)$ be the leakages defined in Definitions \ref{def:fsp-Side-Leakage}, \ref{def:fc-Side-Leakage}, and \ref{def:Bank-Side-Leakage} respectively. If PM is $\epsilon$-differentially private and provides optimal flag privacy (w.r.t. Game presented in Figure \ref{fig:DP-game}), the SecureBoost and $\mathcal{PSI}$ are secure, then \starlit securely realizes \func,  w.r.t. Definition \ref{def::cel-def}. 
\end{theorem}

We prove the above theorem in Appendix \ref{sec::Proof-of-Security}.

% \subsubsection{Mechanisms for Flag Protection: Stackelberg Games and Local Differential Privacy}
%\newpage

%\newpage % only for ease when writing, to be removed
% \input{aggregator}
% \input{model-training}
%\newpage % only for ease when writing, to be removed

%\vspace{-1mm}

\section{Implementation of Starlit }
We carry out comprehensive evaluations to study \starlit's performance from various aspects, including privacy-utility trade-off, efficiency, scalability, and choice of parameters. In the remainder of this section, we elaborate on the analysis. 

%\vspace{-1mm}
\subsection{The Experiment's Environment}

We implement \starlit within an FL framework, called \flower (discussed in Section \ref{sec:flower}). We use Python programming language to implement \starlit. Experiments were run using AWS ECS cloud with docker containers with 56GB RAM and 8 Virtual CPUs. The FATE SecureBoost implementation uses multiprocessing to operate on table-like objects. We set the partitions setting to 5, which means operations on tables are performed with a parallelism of 5.

We adjusted and used the Python-based implementation of the efficient PSI introduced in \cite{KolesnikovKRT16}. 
We have run experiments to evaluate the performance of this PSI.

We conducted the experiments when each party's set's cardinality is in the range $[2^{\st 9},2^{\st 19}]$. Briefly, our evaluation indicates that the PSI's runtime increases from 0.84 to 367.93 seconds when the number of elements increases from $2^{\st 9}$ to $2^{\st 19}$. 
Appendix \ref{sec:Further-Discussion-PSI-Implementation} presents further details on the outcome of the evaluation. Each instance of the PSI, for each account, takes as input string: 
$account_{\st number} ||$ $customer_{\st name} || street_{\st name}$ $|| countryCity_{\st zipcode}$.  The output of the PSI is received by the participating bank. 
To implement \starlit, we had to overcome a set of challenges, including the use of Flower and FATE. In Appendix \ref{sec::challenges}, we discuss these challenges in detail and explain how we addressed them.

%We ran the scenario on the provided dev dataset with \fsp and one bank.

%\vspace{-2mm}

\subsection{Dataset}

Our experiment involves the utilization of two synthetic datasets:

\begin{itemize}[leftmargin=4.7mm,label=$\bullet$]

    \item \underline{Dataset 1}: Synthetic dataset that simulates transaction data obtained from the global payment network of \fsp (acting as \aut).
    
    \item \underline{Dataset 2}: Synthetic dataset related to customers (or users), inclusive of their account information and flags, derived from the partner banks (or clients) of \fsp.
\end{itemize}

Furthermore, the sizes of the datasets are as follows. 

\begin{itemize}[leftmargin=4.7mm,label=$\diamond$]
    \item \fsp's training datasets, in total, contain about 4,000,000 rows. 
    
    \item The banks' dataset includes around 500,000 rows. 

    \item \fsp's test dataset comprises about 700,000 rows.
    
\end{itemize}

\subsubsection{Outline of Dataset 1.}

Each row (or sample) in this dataset corresponds to an individual transaction, signifying a payment from a sending bank to a receiving bank. Each transaction encapsulates details such as the originators and beneficiaries, sender and receiving banks, and payment corridor. The dataset spans approximately a month's worth of transactions and involves fifty institutions. 

It contains various fields such as (a) MessageId: a globally unique identifier, (b) Sender: a bank sending the transaction, (c) Receiver: a bank receiving the transaction, and (d) OrderingAccount: an account identifier for the originating ordering entity. Appendix \ref{secc:DB-1-fields} provides detailed explanations of the fields contained in Dataset 1.

\subsubsection{Outline of Dataset 2.} 

The dataset comprises databases from various banks, encompassing information about their customers' accounts, e.g., the flags associated with each account. Initially, the data was unpartitioned, with all the banks' information consolidated into a single table. 

This dataset contains various fields, for instance: (a) Account: an identifier for the account, (b) Name, name of the account, (c) Street: street address associated with the account, (d) CountryCityZip: remaining address details associated with the account, and (e) Flags: enumerated data type indicating potential issues or special features that have been associated with an account. Appendix \ref{secc:DB-2-fields} elaborates on the fields that Dataset 2 contains.

\section{Empirical Results}\label{sec::empirical-results}

% We assessed \starlit using a synthetic dataset given by a major messaging network provider to financial institutions. The dataset consists of about four million rows. 

Our evaluation of \starlit includes various perspectives (a) privacy-utility trade-off, discussed in Section \ref{sec:privacy-utility-ldp}, (b) efficiency and scalability, explored in Section \ref{sec:efficiency-scalability}, and (c) the choice of concrete parameters, covered in Section \ref{sec:parameter-selection}.

% In this section, we discuss the empirical privacy-utility trade-off that can be achieved with \starlit and compare it with a baseline centralised solution without privacy protections. We also discuss the efficiency and scalability aspects of \starlit.  

%\subsubsection{Features Used.} 

In this study, our focus does not lie on feature exploration or hyperparameter tuning to enhance model accuracy. Instead, we employ a straightforward approach, utilizing example features extracted from \fsp, as provided in the data, in conjunction with four binary values derived from the banks' data. 
The features extracted from \fsp for model training encompass the following: settlement amount, instructed amount, hour, sender hour frequency, sender currency frequency, sender currency amount average, and sender-receiver frequency. Additionally, we incorporate four binary flags, indicating the agreement between \fsp and the banks on sender and receiver address details, as well as whether the sending and receiving accounts share the same flag for a given account.

 %\vspace{-2mm}
\subsection{Privacy-Utility Trade-off}\label{sec:privacy-utility-ldp}

\subsubsection{Baseline.}

To analyze the trade-off between utility and privacy, we establish a benchmark using a \textit{centralized} model constructed within \fsp. In this centralized model, all data from banks is revealed in plaintext. The same set of features listed above is extracted. We train a standard XGBoost model with 30 trees. We employ a 5-fold cross-validation with the average precision score as the metric. It is important to note that default values are utilized for all hyperparameters during the model training process.

\subsubsection{Evaluation Procedure.} 

 The ``Area Under the Precision-Recall Curve'' (AUPRC) refers to a metric employed to evaluate the performance of an ML classification model. The unit of AUPRC is a value in the range $[0, 1]$, representing the area under the precision-recall curve. It measures the trade-off between precision and recall and provides a summary of the model's performance across different threshold values for classification. A higher AUPRC indicates better model performance, with 1 being the ideal value representing perfect precision and recall.

\subsubsection{Starlit.} In the evaluation of \starlit's implementation, for analyzing AUPRC that can be achieved at a given level of privacy, we modify the flag values that banks send using DP and construct XGBoost models with these noisy features. 

We use the same parameters as in the baseline model (30 trees and 5-fold cross-validation) and measure the average precision score for the final model on training and test data, averaging over 5 runs to account for the randomness of the privacy mechanism and the training process.

%We chose XGBoost for the baseline and also in our experiments to understand the privacy-utility trade-off. %as it can be theoretically proved that it will achieve the same accuracy as our federated SecureBoost solution. 
%
% , given the same set of parameters and input data. 
SecureBoost does the same computation as XGBoost while constructing the trees albeit on encrypted gradients. Hence, the additional cost will not be on accuracy but rather on performance (which we discuss in section~\ref{sec:efficiency-scalability}). We also observed this to be the case from our experimental results.

\begin{figure}[!h]
  \begin{subfigure}{0.235\textwidth}
   %%\vspace{-5mm}
  \centering
    \includegraphics[width=\linewidth]{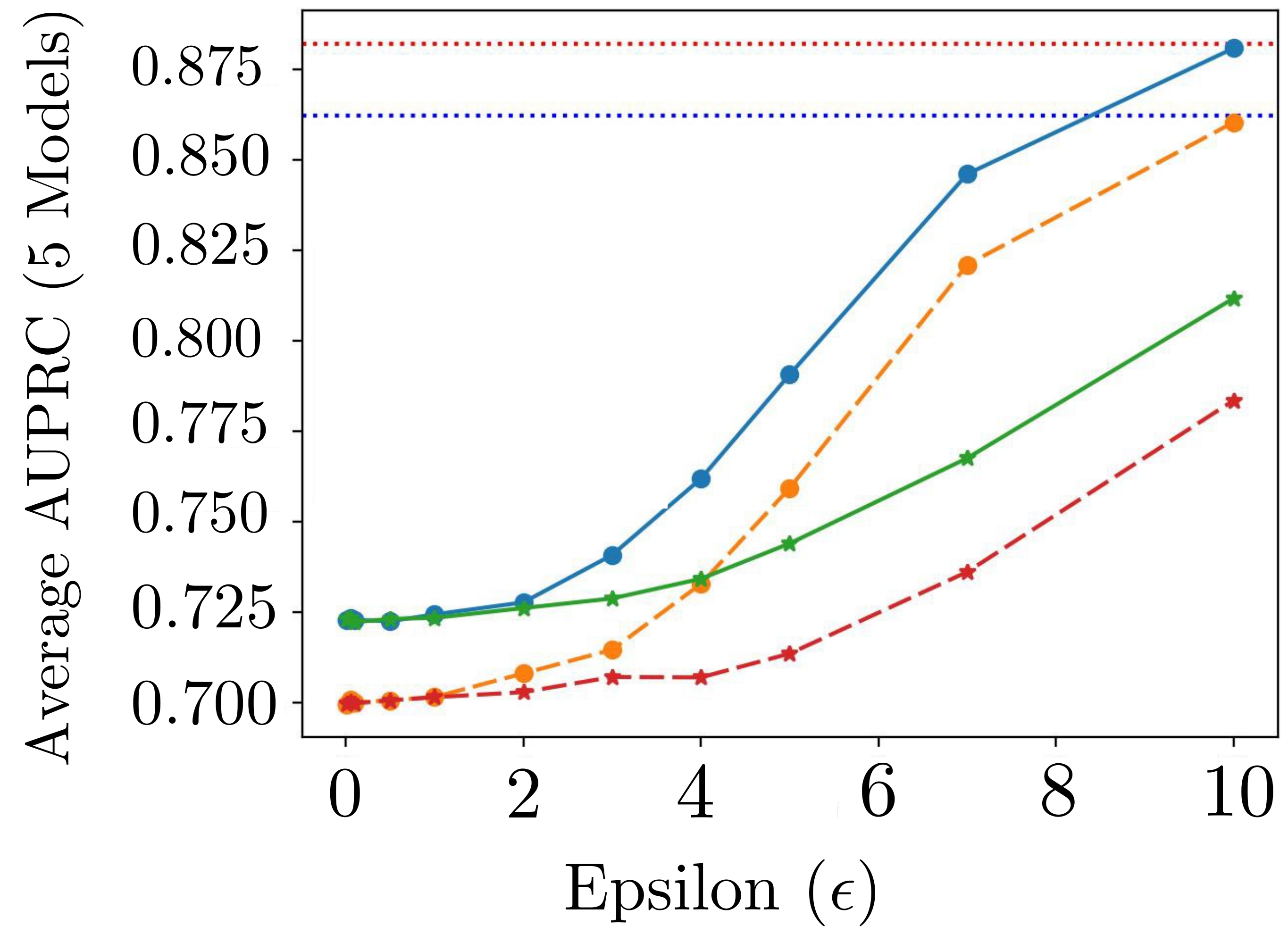}
    %%\vspace{-5mm}
    \captionsetup{font=small}
    \caption{\scriptsize{Rand. Response vs Laplace. }}\label{fig:rr-laplace}
  \end{subfigure}%
  \begin{subfigure}{0.235\textwidth}%{0.45\textwidth}
    \includegraphics[width=\linewidth]{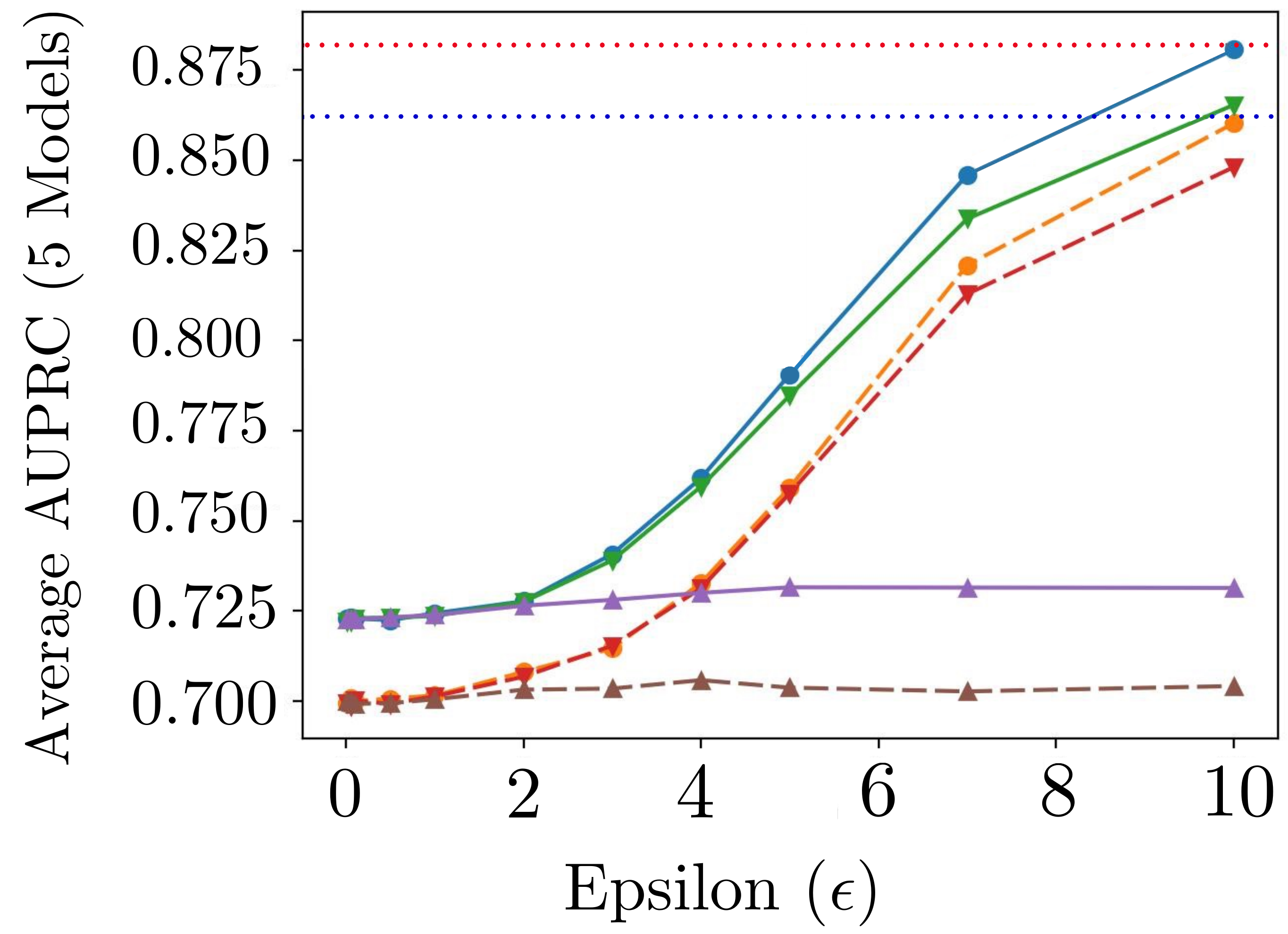}
    %\vspace{-4mm}
    \captionsetup{font=small}
    \caption{\scriptsize{Rand. Response vs Asym. matrices.  
    }}\label{fig:game-rr}
  \end{subfigure}
  \captionsetup{font=small}
  \caption{\small{Plot(\protect\subref{fig:rr-laplace}) compares the effect on AUPRC of the model when using RR and Laplace mechanism with post-processing for achieving LDP. Plot(\protect\subref{fig:game-rr}) compares the effect on AUPRC when using RR and privacy mechanisms at the same value of $\epsilon$ but with the constraint of 10\% less probability of converting 0 to 1 (1 to 0) than what is recommended by RR.
  In Plot(\protect\subref{fig:rr-laplace}), red dotted line: non-private-train,  blue dotted line: non-private-test, solid blue line: RR-train, solid orange line: RR-test, solid green line: Laplace-train, and solid red line: Laplace-test. 
  In Plot(\protect\subref{fig:game-rr}), red dotted line: non-private-train,  blue dotted line: non-private-test, solid blue line: RR-train, solid orange line: RR-test, solid green line: 10\% less 0->1 than RR-train, 
    solid red line: 10\% less 0->1 than RR-test, 
    solid purple line: 10\% less 1->0 than RR-train,
    and solid brown line: 10\% less 1->0 than RR-test.
  }}\label{fig:results-plots}
  %\vspace{-1mm}
\end{figure}

\subsubsection{Key Takeaways.} 

Figure \ref{fig:results-plots} provides a summary of our utility-privacy trade-off analysis. 
Plot(\protect\subref{fig:rr-laplace}) in this figure compares the effect on AUPRC of the model when using Randomized Response (RR) and Laplace mechanism with post-processing for achieving LDP. 
Consistent with the optimality results presented in \cite{wang2016using,kairouz2014extremal}, RR offers a superior utility-privacy trade-off when compared to the Laplace mechanism. Both RR and the Laplace mechanism yield symmetric transformation matrices, meaning an equal probability for converting a 0 to 1 and a 1 to 0.

Plot(\protect\subref{fig:game-rr}) in Figure \ref{fig:results-plots}
 illustrates the impact on AUPRC when employing RR and privacy mechanisms. This comparison is conducted at the same $\epsilon$ value, with the additional constraint of reducing the probability of converting 0 to 1 (and 1 to 0) by 10\% compared to the recommendations provided by RR. These recommendations are determined using our game framework. 
The results demonstrate that even a slight increase in the probability of converting a zero flag to a non-zero value has a significant impact on the model's performance. This observation aligns with intuition, considering the substantial proportion of zero flag values in the dataset.

\subsection{Efficiency and Scalability} \label{sec:efficiency-scalability}

% Experiments were run using AWS ECS cloud with the docker environment provided for the competition. The docker containers are provisioned with 56GB RAM and 8 vcpu. The FATE secureboost implementation uses multiprocessing to operate on table-like objects. We set the partitions setting to 5, which means operations on tables are performed with a parallelism of 5.
% We ran the scenario on the provided dev dataset with \fsp and one bank.

\subsubsection{Baseline.}
%Our solution can be split into two main phases: feature extraction and training (or prediction). 
%
%In the baseline (centralised) setting, \fsp does not have access to information from other parties. Consequently, in this scenario, \fsp undertakes the steps involved in the second phase, which is training. 
%

SecureBoost's training was configured with $10$ trees, each with a depth of $3$, a dataset sampling rate of $40\%$, and a ``Gradient-based One Side Sampling'' (GOSS) sampling of $0.1$. Efficiency results for this baseline are provided in Table \ref{table::efficiency-baseline-table}. This baseline is used to investigate various configurations' impact on efficiency.

% \begin{table*}[!htbp]
% \begin{center}
% \caption{\small Efficiency metrics for baseline performance investigation.}\label{table::efficiency-baseline-table}

% \begin{tabular}{|l|l|l|}
% \hline
% AUPRC                                           & 0.4715 \\ \hline
% total\_training\_time\_full-data-federated (in hour) & 1.1    \\ \hline
% peak\_training\_memory\_full-data-federated (in GB) & 12.38  \\ \hline
% network\_disk\_volume\_full-data-federated (in GB)  & 4.98   \\ \hline
% network\_file\_volume\_full-data-federated (in GB)     & 993    \\ \hline
% \end{tabular}

% \end{center}
% \end{table*}

%%%%%%%%%
\begin{table}[!h]
%\begin{footnotesize}
%\begin{center}
%\vspace{-3mm}
\caption{Efficiency metrics of the baseline. H represents time in hours and GB refers to gigabytes.}  \label{table::efficiency-baseline-table}

\begin{center}
\renewcommand{\arraystretch}{1}
\scalebox{1}{
\begin{tabular}{|c|c|c|c|c|c|c|c|c|c|c|c|} 
    \cline{1-4}

 \cellcolor{\clr}\scriptsize Efficiency Metric&\cellcolor{\clr} \scriptsize Unit&\cellcolor{\clr} \scriptsize Tree's depth &\cellcolor{\clr} \scriptsize Result\\

       \hline
       
\cellcolor{white!20}\scriptsize AUPRC&\scriptsize --& \scriptsize 3 &\cellcolor{gray!50}\scriptsize   0.4715\\
    
        \hline

\cellcolor{white!20}\scriptsize The total training time&\scriptsize  H& \scriptsize 3&\cellcolor{gray!20}\scriptsize   1.1\\ 

        \hline
     
\cellcolor{white!20}\scriptsize The peak training memory usage&\scriptsize GB & \scriptsize 3&\cellcolor{gray!50}\scriptsize  12.38\\ 

       \hline
    
\cellcolor{white!20}\scriptsize The network disk volume usage&\scriptsize  GB& \scriptsize 3&\cellcolor{gray!20}\scriptsize   4.98\\ 

    \hline

\cellcolor{white!20}\scriptsize The network file volume usage&\scriptsize  GB& \scriptsize 3&\cellcolor{gray!50}\scriptsize   993\\ 

    \hline

\end{tabular}
}
\end{center}
%\end{footnotesize}
%\vspace{-3mm}
\end{table}

%%%%%%%%%

\subsubsection{Starlit.} We analyzed \starlit's efficiency with different SecureBoost configurations. The evaluation's results are illustrated in Table \ref{table::efficiency-table}. SecureBoost offers various options that can be employed to enhance efficiency in different settings. For instance, both direct sampling\footnote{This direct sampling approach is taken to reduce the training time. It works by randomly using only the stated fraction of the training set.} and GOSS sampling offers a means to reduce network and memory overhead by decreasing the volume of data processed in each round of training. 

The tree depth is also a crucial parameter for improving accuracy while maintaining an appropriate level of efficiency in terms of training time and memory consumption. 
Also, the integration of \starlit with FATE and \flower enables the splitting of large messages into chunks, facilitating more efficient processing. 
Furthermore, \starlit utilizes numerous \flower rounds, with a significant portion of the final rounds remaining empty due to the requirement of a pre-set round number by \flower. This situation has an impact on the network metrics.

\begin{table}[!htb]
%\begin{footnotesize}
\begin{center}
%\vspace{-3mm}
\caption{ \small  \starlit's Runtime using various training parameters. In the table, H represents time in hours and GB refers to gigabyte. The row highlighted in yellow corresponds to the choice of parameters where AUPRC is at the highest level.}\label{table::efficiency-table} 
%\vspace{-.5mm}

\renewcommand{\arraystretch}{.9}
\scalebox{0.8}{
\begin{tabular}{|c|c|c|c|c|c|c|c|c|c|c|c|c|c|c|} 
%\cline{1-9}
   \hline
\cellcolor{\clr}&\cellcolor{\clr}&
 \multicolumn{2}{c|}{\cellcolor{\clr}\scriptsize Sampling Approach}&\multirow{2}{*} {\cellcolor{\clr}\scriptsize Tree's depth} &\cellcolor{\clr}&\cellcolor{\clr}\\

\cline{3-4}

 \multirow{-2}{*} {\cellcolor{\clr}\scriptsize Efficiency Metric}  &\multirow{-2}{*} {\cellcolor{\clr}\scriptsize Unit}&\cellcolor{\clr}\scriptsize{Direct Sampling Rate} &\cellcolor{\clr}\scriptsize{GOSS}&\multirow{-2}{*} {\cellcolor{\clr}\scriptsize Tree's depth} & \multirow{-2}{*} {\cellcolor{\clr}\scriptsize Max Message Size}&\multirow{-2}{*} {\cellcolor{\clr}\scriptsize Result} \\
            
 \cline{1-7}

%%%%%%%%%%%%%%%%%%%%
 \multirow{6}{*}{\rotatebox[origin=c]{35}{\scriptsize  AUPRC }}& \multirow{6}{*}{\rotatebox[origin=c]{0}{\  \scriptsize -- }}&\scriptsize 
 40\%&\scriptsize 0.1&\scriptsize 3 &\scriptsize 100MB&\cellcolor{gray!20}\scriptsize  0.4715 \\

     \cline{3-7}  

     &&\scriptsize 100\%& \scriptsize 0.1&\scriptsize 3&\scriptsize 100MB&\cellcolor{gray!50}\scriptsize  0.5786\\
     \cline{3-7} 

         &&\scriptsize 40\%& \scriptsize Disabled&\scriptsize 3 &\scriptsize 100MB&\cellcolor{gray!20}\scriptsize 0.47 \\

     \cline{3-7} 

         &&\scriptsize 40\%& \scriptsize 0.3&\scriptsize 3 &\scriptsize 100MB&\cellcolor{gray!50}\scriptsize 0.5965 \\

     \cline{3-7} 

         &&\cellcolor{yellow!20}\scriptsize 40\%&\cellcolor{yellow!20} \scriptsize 0.1&\cellcolor{yellow!20}\scriptsize 5&\cellcolor{yellow!20}\scriptsize 100MB&\cellcolor{yellow!20}\scriptsize 0.652 \\

              \cline{3-7} 

         &&\scriptsize 40\%& \scriptsize 0.1&\scriptsize 3 &\scriptsize 1GB&\cellcolor{gray!50}\scriptsize 0.4715 \\

      \cline{1-7}

      \cline{1-7}
%%%%%%%%%%%%

%%%%%%%%%%%%%%%%%%%%
 \multirow{6}{*}{\rotatebox[origin=c]{35}{\scriptsize  The total training time }}& \multirow{6}{*}{\rotatebox[origin=c]{0}{\  \scriptsize H }}&\scriptsize 
 40\%&\scriptsize 0.1&\scriptsize 3 &\scriptsize 100MB&\cellcolor{gray!20}\scriptsize 1.1 \\

     \cline{3-7}  

     &&\scriptsize 100\%& \scriptsize 0.1&\scriptsize 3&\scriptsize 100MB&\cellcolor{gray!50}\scriptsize  2.21\\
     \cline{3-7} 

         &&\scriptsize 40\%& \scriptsize Disabled&\scriptsize 3 &\scriptsize 100MB&\cellcolor{gray!20}\scriptsize 2.83 \\

     \cline{3-7} 

         &&\scriptsize 40\%& \scriptsize 0.3&\scriptsize 3 &\scriptsize 100MB&\cellcolor{gray!50}\scriptsize  1.5 \\

     \cline{3-7} 

         &&\scriptsize 40\%& \scriptsize 0.1&\scriptsize 5&\scriptsize 100MB&\cellcolor{gray!20}\scriptsize 1.13 \\

              \cline{3-7} 

         &&\scriptsize 40\%& \scriptsize 0.1&\scriptsize 3 &\scriptsize 1GB&\cellcolor{gray!50}\scriptsize 1 \\

  %%%%%%%%%%%%%%%%%%%%

\cline{1-7}
  
 \multirow{6}{*}{\rotatebox[origin=c]{35}{\scriptsize  The peak training  }}& \multirow{6}{*}{\rotatebox[origin=c]{0}{\scriptsize GB }}&\scriptsize 
 40\%&\scriptsize 0.1&\scriptsize 3 &\scriptsize 100MB&\cellcolor{gray!20}\scriptsize 12.38 \\

     \cline{3-7}  

      \multirow{6}{*}{\rotatebox[origin=c]{35}{\scriptsize memory usage}}      &&\scriptsize 100\%& \scriptsize 0.1&\scriptsize 3&\scriptsize 100MB&\cellcolor{gray!50}\scriptsize  17.48\\
     \cline{3-7} 

   &&\scriptsize 40\%& \scriptsize Disabled&\scriptsize 3 &\scriptsize 100MB&\cellcolor{gray!20}\scriptsize 18.39\\

     \cline{3-7} 

         &&\scriptsize 40\%& \scriptsize 0.3&\scriptsize 3 &\scriptsize 100MB&\cellcolor{gray!50}\scriptsize  13.66\\

     \cline{3-7} 

         &&\scriptsize 40\%& \scriptsize 0.1&\scriptsize 5&\scriptsize 100MB&\cellcolor{gray!20}\scriptsize 16.4 \\

              \cline{3-7} 

         &&\scriptsize 40\%& \scriptsize 0.1&\scriptsize 3 &\scriptsize 1GB&\cellcolor{gray!50}\scriptsize 12.22\\

  %%%%%%%%%%%%%%%%%%%%

\cline{1-7}
  
 \multirow{6}{*}{\rotatebox[origin=c]{35}{\scriptsize  The network disk   }}& \multirow{6}{*}{\rotatebox[origin=c]{0}{\scriptsize GB }}&\scriptsize 
 40\%&\scriptsize 0.1&\scriptsize 3 &\scriptsize 100MB&\cellcolor{gray!20}\scriptsize 4.98 \\

     \cline{3-7}  

      \multirow{6}{*}{\rotatebox[origin=c]{35}{\scriptsize volume usage}}      &&\scriptsize 100\%& \scriptsize 0.1&\scriptsize 3&\scriptsize 100MB&\cellcolor{gray!50}\scriptsize  14.51\\
     \cline{3-7} 

   &&\scriptsize 40\%& \scriptsize Disabled&\scriptsize 3 &\scriptsize 100MB&\cellcolor{gray!20}\scriptsize 16.61\\

     \cline{3-7} 

         &&\scriptsize 40\%& \scriptsize 0.3&\scriptsize 3 &\scriptsize 100MB&\cellcolor{gray!50}\scriptsize  7.84\\

     \cline{3-7} 

         &&\scriptsize 40\%& \scriptsize 0.1&\scriptsize 5&\scriptsize 100MB&\cellcolor{gray!20}\scriptsize 5.1 \\

              \cline{3-7} 

         &&\scriptsize 40\%& \scriptsize 0.1&\scriptsize 3 &\scriptsize 1GB&\cellcolor{gray!50}\scriptsize 4.34 \\

  %%%%%%%%%%%%%%%%%%%%

\cline{1-7}
  
 \multirow{6}{*}{\rotatebox[origin=c]{35}{\scriptsize     The network file}}& \multirow{6}{*}{\rotatebox[origin=c]{0}{\scriptsize GB }}&\scriptsize 
 40\%&\scriptsize 0.1&\scriptsize 3 &\scriptsize 100MB&\cellcolor{gray!20}\scriptsize 993 \\

     \cline{3-7}  

      \multirow{6}{*}{\rotatebox[origin=c]{35}{\scriptsize volume usage}}      &&\scriptsize 100\%& \scriptsize 0.1&\scriptsize 3&\scriptsize 100MB&\cellcolor{gray!50}\scriptsize  1270 \\
     \cline{3-7} 

   &&\scriptsize 40\%& \scriptsize Disabled&\scriptsize 3 &\scriptsize 100MB&\cellcolor{gray!20}\scriptsize 1256 \\

     \cline{3-7} 

         &&\scriptsize 40\%& \scriptsize 0.3&\scriptsize 3 &\scriptsize 100MB&\cellcolor{gray!50}\scriptsize 1035 \\

     \cline{3-7} 

         &&\scriptsize 40\%& \scriptsize 0.1&\scriptsize 5&\scriptsize 100MB&\cellcolor{gray!20}\scriptsize 1316  \\

              \cline{3-7} 

         &&\scriptsize 40\%& \scriptsize 0.1&\scriptsize 3 &\scriptsize 1GB&\cellcolor{gray!50}\scriptsize 927 \\

%%%%%%%%%%%
 \hline
\end{tabular}
}
\end{center}
%\end{footnotesize}
\vspace{2mm}
\end{table}

%%%%%%%%%%%%%

%\subsubsection{Scalability.} 

Furthermore, \starlit's scalability is attributed to its design choice, which maintains the model training phase's independence from the number of banks. In contrast, the feature extraction phase's computational complexity scales linearly with the number of banks and the suspicious accounts identified by \fsp.

% Also, \starlit is \textit{highly scalable}. This is due to the \starlit design choice that keeps the model training phase independent of the number of banks. Instead,  the training phase involves only \fsp and \fc.  On the other hand, the computation complexity of the feature extraction phase scales linearly with the number of banks and the number of accounts that seem suspicious from \fsp's perspective. 

%The feature extraction scales with the number of bank accounts involved rather than the number of banks, aside from a small overhead due to additional Flower communication. %Table \ref{table::scalability-table} demonstrates our experimental results confirming this.

 % \begin{table*}[!htbp]

% \begin{center}
% \caption{\small The runtime of feature extraction depending of the number of bank clients}\label{table::scalability-table} 

% \begin{tabular}{|l|l|l|l|}
% \hline
% number of bank clients               & 2                           & 5                           & 9                           \\
% \hline
% feature extraction runtime (seconds) & \cellcolor{gray!20}833 & \cellcolor{gray!20}808 & \cellcolor{gray!20}821 \\
% \hline
% \end{tabular}

% \end{center}
% \end{table*}

\starlit extensively employs CPU resources, primarily driven by the CPU-intensive nature of the underlying SecureBoost training. This demand arises from the encryption necessary for securing the gradient in the algorithm. Refer to Figure \ref{fig:cpu-baseline} in Appendix \ref{sec::CPU-Utilization} for the CPU utilization details of \starlit.

% \starlit utilizes a high level of CPU, as the underlying SecureBoost training is CPU intensive, this is due to the encryption required by the algorithm to encrypt the gradient. The CPU utilization of \starlit is provided in Figure \ref{fig:cpu-baseline} in Appendix \ref{sec::CPU-Utilization}. %In the figure, we used 5 CPUs and did not use any CPU-specific library to run Paillier encryption, used as a subroutine in SecureBoost. %The time that takes to train or predict can be reduced by increasing parallelism or using hardware-specific libraries like the Intel Paillier Cryptosystem Library. 

 \subsection{Choice of Parameters} \label{sec:parameter-selection}

Our ultimate selection of parameters was informed by (i) the privacy-utility analysis in Section \ref{sec:privacy-utility-ldp}, specifically in the selection of $\epsilon$ and (ii) the efficiency analysis in Section \ref{sec:efficiency-scalability}, pertaining to the determination of model hyper-parameters, such as AUPRC, the number of trees, and sampling rate. 

For the centralized solution, as far as possible, we matched the parameters used in the centralized XGBoost model with those used in the federated solution. Specifically for the number of trees, tree depth, and L2 regularization parameter. The parameters are set as follows: (a) in both the centralized and federated settings: number of trees = 10, L2 regularization = 0.1, and (b)  in the federated setting: $\epsilon = 10$.

% The parameters selected are as presented in Table \ref{table::sys-parameters}.

% follows:

% \begin{itemize}[label = $\bullet$]

% \item \textbf{Centralised:} $number\_of\_trees=10, tree\_depth=5, L2\_regularisation=0.1$

% \item \textbf{Federated:} $number\_of\_trees=10, tree\_depth=5, L2\_regularisation=0.1, epsilon=10$

% \end{itemize}

% %%%%%%%%%
% \begin{table}[!h]
% %\begin{footnotesize}
% %\begin{center}
% %\vspace{-4mm}
% \caption{The choice of parameters for centralized and federated settings.}  \label{table::sys-parameters}
% %\vspace{-1mm}
% \begin{center}
% \renewcommand{\arraystretch}{1}

% \begin{tabular}{|c|c|c|c|c|c|c|c|c|c|c|c|} 
%     \cline{1-5}

%  \cellcolor{\clr}\scriptsize Setting&\cellcolor{\clr} \scriptsize Number of trees&\cellcolor{\clr} \scriptsize Tree depth&\cellcolor{\clr} \scriptsize L2 regularization&\cellcolor{\clr} \scriptsize $\epsilon$\\

%        \hline
       
% \cellcolor{white!20}\scriptsize Centralized&\cellcolor{gray!20}\scriptsize 10  &\cellcolor{gray!20}\scriptsize   5&\cellcolor{gray!20}\scriptsize  0.1&\cellcolor{gray!20}\scriptsize -- \\
    
%         \hline

% \cellcolor{white!20}\scriptsize Federated&\cellcolor{gray!20}\scriptsize 10  &\cellcolor{gray!20}\scriptsize   5&\cellcolor{gray!20}\scriptsize  0.1&\cellcolor{gray!20}\scriptsize 10 \\

%     \hline
    
% \end{tabular}
% \end{center}
% %\end{footnotesize}
% %\vspace{-1mm}
% \end{table}

% %%%%%%%%%

 %\vspace{-2mm}
\subsection{Contrasting Starlit with the Baseline} 

In this section, we provide a brief comparison of \starlit's performance with the baseline scenario, drawing insights from the information presented in Tables \ref{table::efficiency-baseline-table} and \ref{table::efficiency-table}. 

\begin{figure}[H]
    \centering
    \includegraphics[width=6cm]{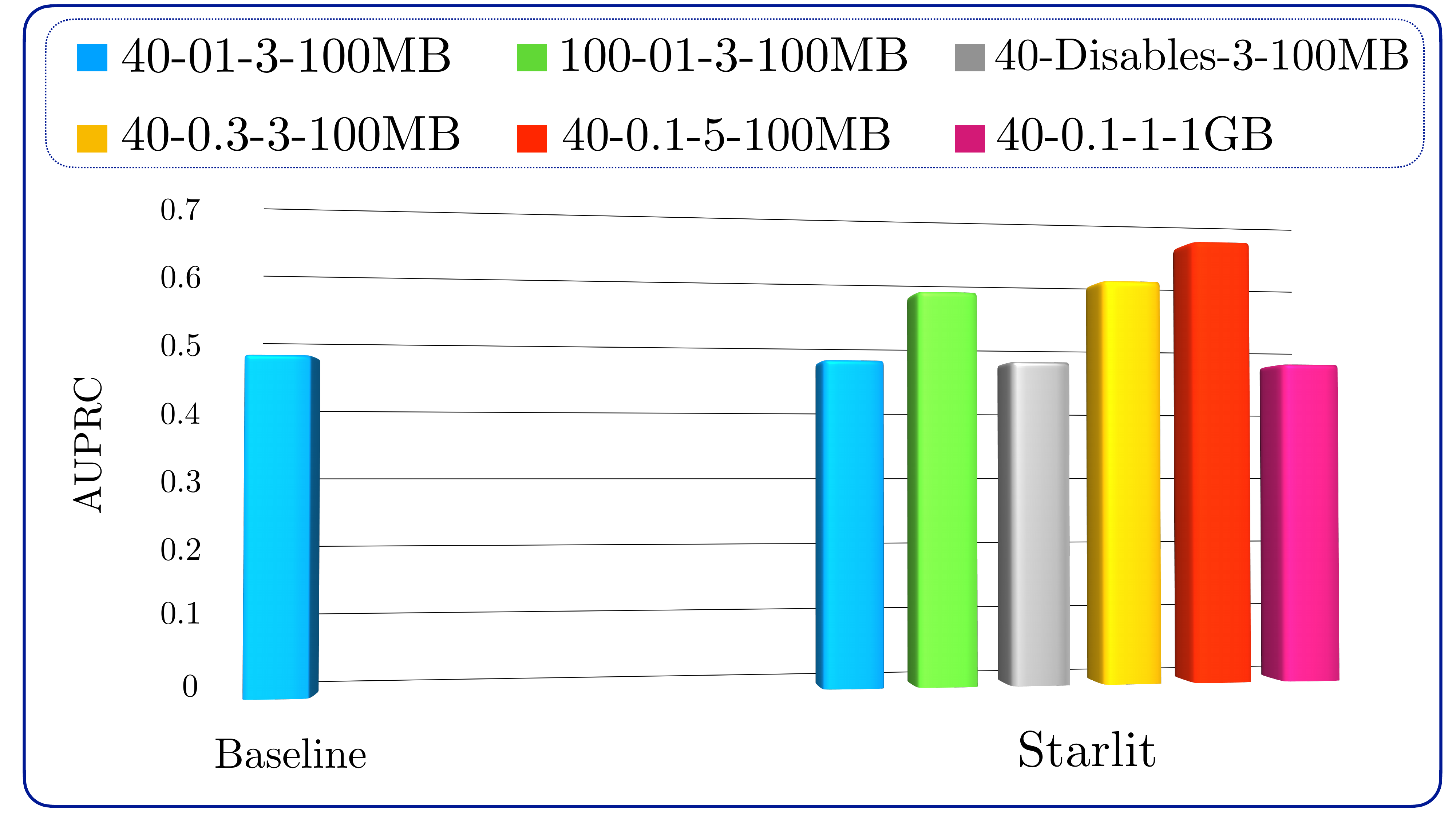}
    %%\vspace{-1mm}
         \caption{Comparing the AUPRC of the baseline and different settings of \starlit. A bar's label for \starlit is a concatenation of (1)  direct sampling rate, (2) GOSS, (3) tree's depth, and (4) maximum message size.}\label{fig:bar-char-AUPRC} 
  %\vspace{-3mm}
\end{figure}

 \starlit and the baseline achieve the same level of AUPRC when \starlit's (i) direct sampling rate is $40\%$, (ii) the tree's depth is $3$, and (iii) GOSS is not disabled. This is indicated in Figure \ref{fig:bar-char-AUPRC}. In the case where direct sampling rate = $40\%$ and tree's depth = $3$, regarding:

\

\begin{itemize}[leftmargin=4mm, label=$\diamond$]
    \item \underline{\textit{The total training time}}: \starlit and the baseline have similar performance, except for the case when GOSS in \starlit is disabled. In this case, \starlit underperforms the baseline by a factor of $2.5$.

    \item \underline{\textit{The peak training memory usage}}: \starlit and the baseline have similar performance except for the cases where (i) GOSS in \starlit is disabled and (ii) GOSS is $0.3$. In these cases, \starlit underperforms the baseline by at most a factor of  $1.4$. 
    
    \item \underline{\textit{The network disk volume usage}}: \starlit and the baseline have similar performance when GOSS = $0.1$ and max message size = $100$ MB. However, when max message size = $1$ GB, \starlit outperforms the baseline by a factor of $1.1$. In the rest of the cases, \starlit underperforms the baseline by at most a factor of $3.3$.

    \item \underline{\textit{The network file volume usage}}: \starlit and the baseline have similar performance when direct sampling rate = $40\%$, GOSS = $0.1$, and max message size = $100$ MB. However, when max message size = $1$ GB, \starlit outperforms the baseline by a factor of $1.07$. In the rest of the cases, the baseline outperforms \starlit by at most a factor of $1.27$.
\end{itemize}

  %\vspace{-2mm}

As illustrated in Table \ref{table::efficiency-table}, \starlit achieves its highest AUPRC level (i.e., $0.652$) when the tree's depth is set to $5$. Remarkably, in this instance, \starlit's AUPRC surpasses even the baseline setting (i.e., $0.652$ versus $0.4715$). 
%
% The key factor contributing to this difference is 
% the tree's depth. 

Having identified the parameter that yields the highest AUPRC in \starlit, i.e., when the tree's depth is set to $5$, we proceed to compare \starlit's other efficiency metrics for only that parameter(s). 

\begin{figure*}[h]
    \centering
    \renewcommand{\arraystretch}{.02}
    \begin{subfigure}{0.234\textwidth}
        \includegraphics[width=\linewidth]{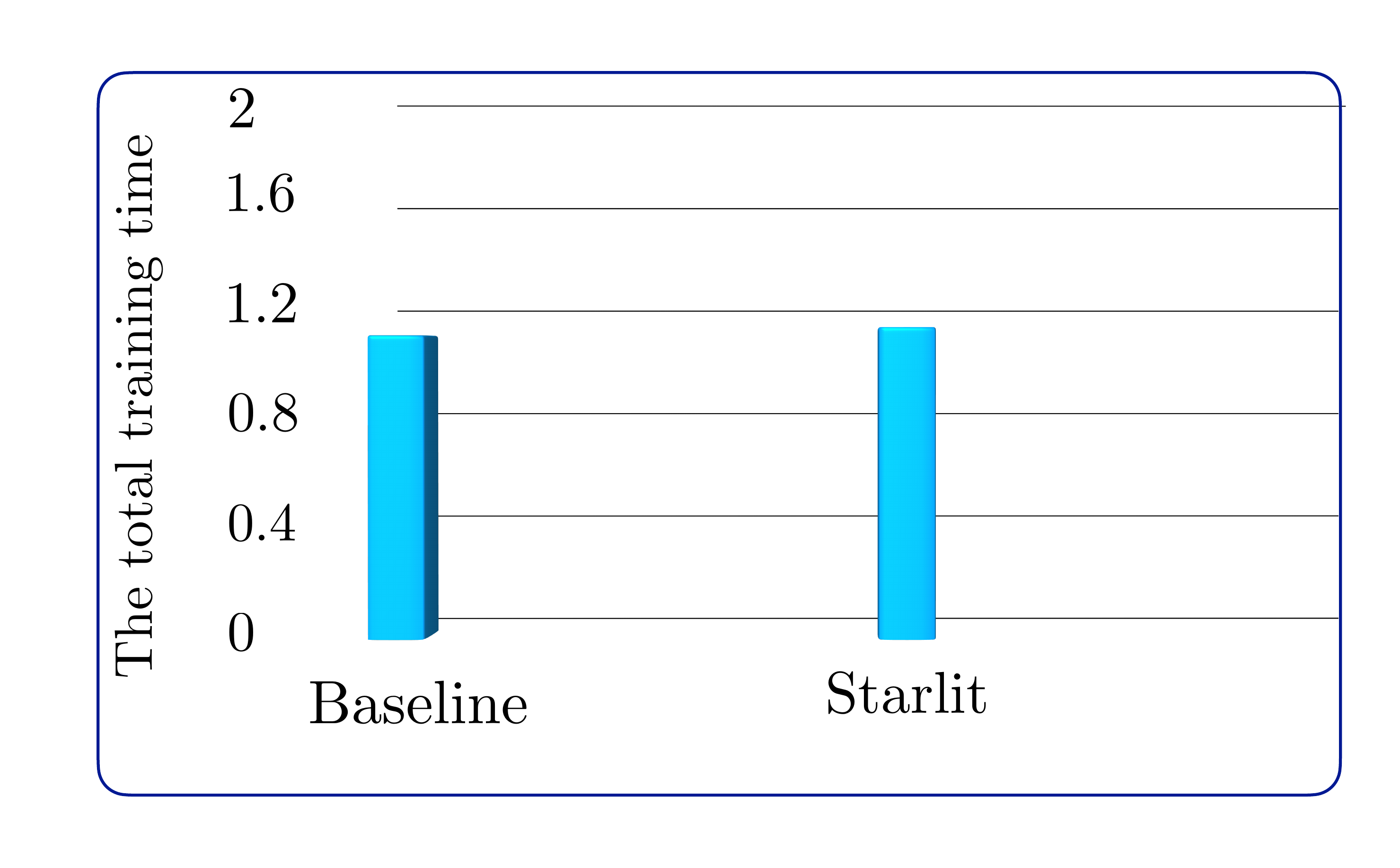}
        %\input{training-runtime-comparison-tree-depth-5}
        %}
        %\vspace{-2mm}
        \caption{{\small{Comparison between the total training time of the baseline and \starlit.}}}\label{fig::run-time-B-vs-S5}
    \end{subfigure}
    %
   %\hfill
    \begin{subfigure}{0.234\textwidth}
    \includegraphics[width=\linewidth]{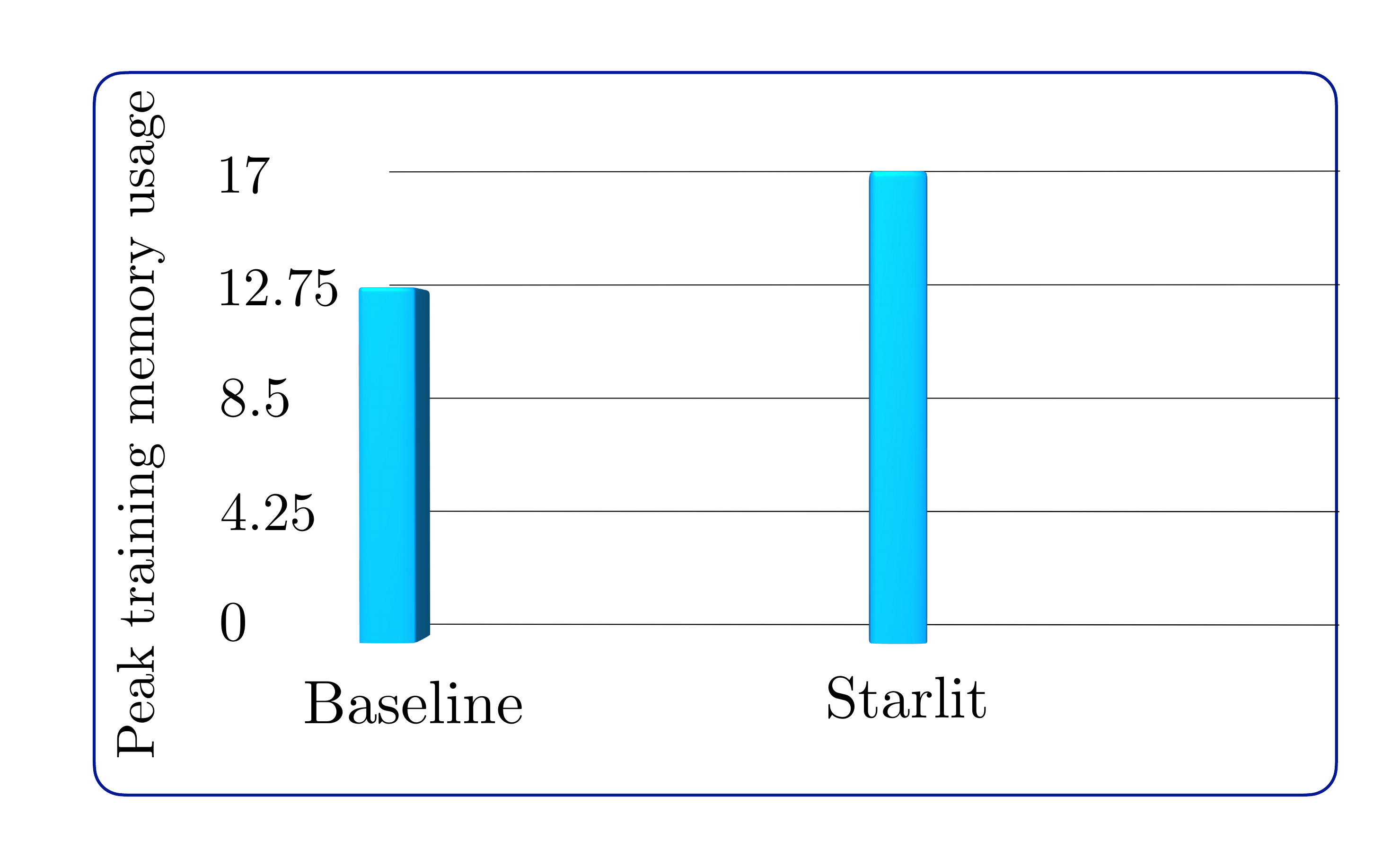}
        \caption{{\small{Comparison between peak training memory usage of the baseline and \starlit.}}}\label{fig::peack-training-B-vs-S5}
    \end{subfigure}
    %
    %\vspace{10pt} % Adjust the vertical space between rows
    \begin{subfigure}{0.234\textwidth}
     \includegraphics[width=\linewidth]{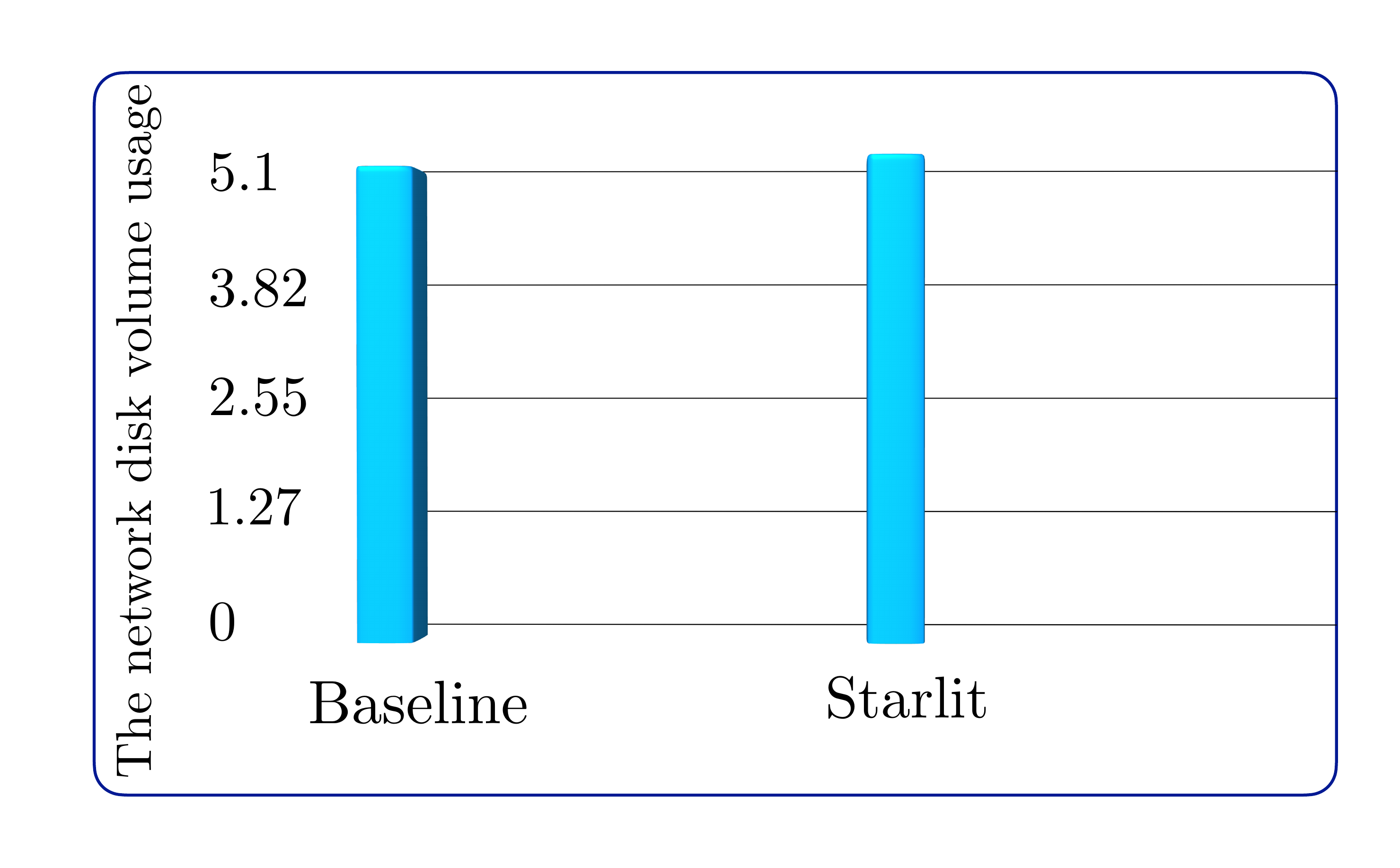}
        %\includegraphics[width=\linewidth]{
        %\input{network-disk-volume-usage}
        %}
       % \vspace{-2mm}
        \caption{Comparison between network disk volume usage of the baseline and \starlit.}\label{fig::network-disk-volume-B-vs-S5}
    \end{subfigure}
%
    %\hfill
    \begin{subfigure}{0.234\textwidth}
     \includegraphics[width=\linewidth]{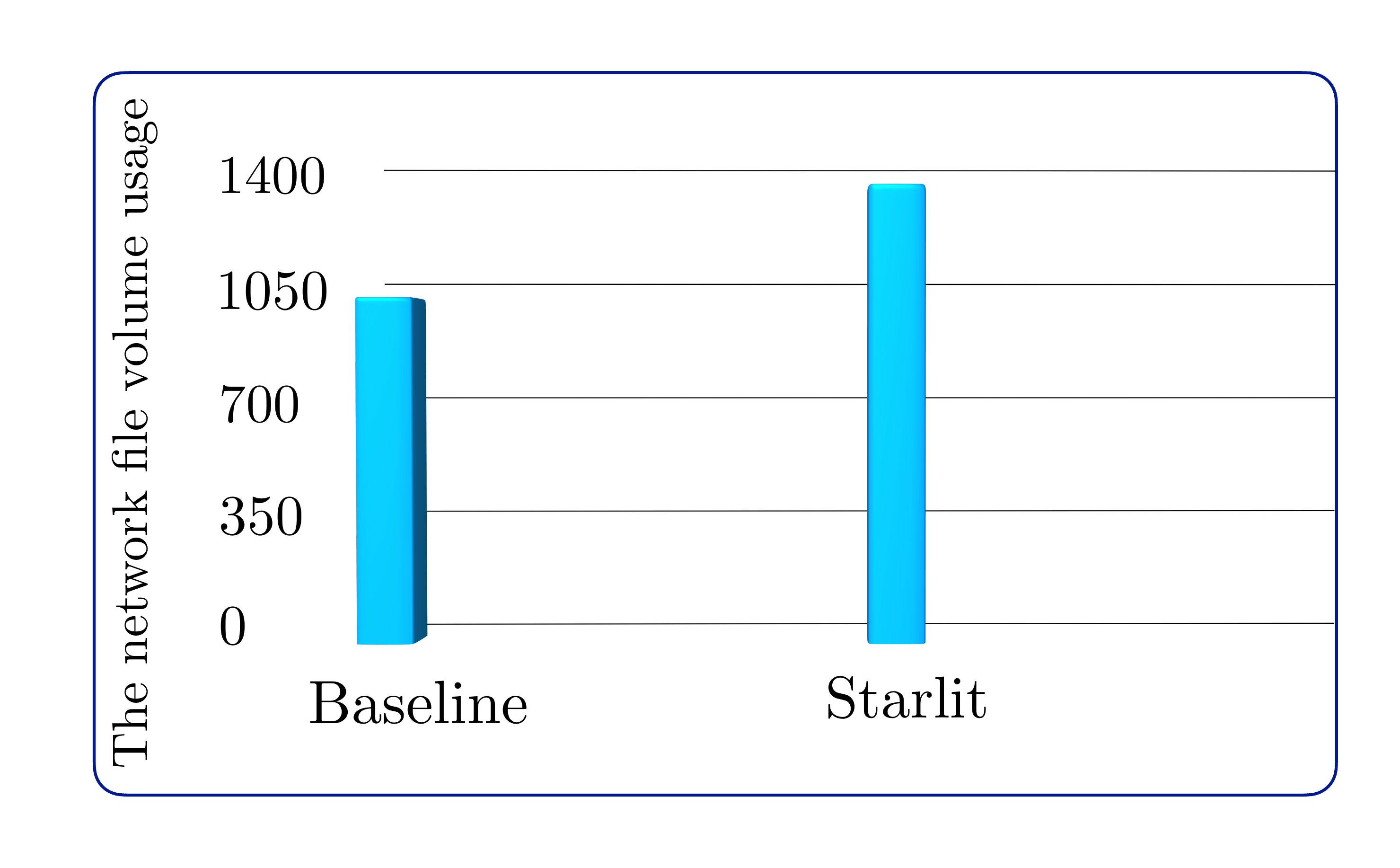}
       % \includegraphics[width=\linewidth]{
        %\input{network-file-volume-usage}
       % }
       %\vspace{-2mm}
        \caption{Comparison between network file volume usage of the baseline and \starlit.}\label{fig::network-file-volume-B-vs-S5}
    \end{subfigure}
\vspace{-1mm}
    \caption{In this figure,  \starlit's: sampling rate = 40, GOSS = 0.1, tree's depth = 5, and max message size = 100MB.}
    %\vspace{-1mm}
\end{figure*}

\begin{itemize}[leftmargin=4mm,label=$\bullet$]

   % \item \underline{\textit{The total training time}}: As Figure \ref{fig::run-time-B-vs-S5} demonstrates, \starlit slightly underperforms the baseline, i.e., $1.13$ compared to $1.10$. 

    \item \underline{\textit{The total training time}}: As Figure \ref{fig::run-time-B-vs-S5} demonstrates, \starlit's training time is almost the same as the baseline's training time, i.e., $1.13$ compared to $1.10$. 
    
    \item  \underline{\textit{The peak training memory usage}}: As Figure \ref{fig::peack-training-B-vs-S5} illustrates, under this metric, \starlit underperforms the baseline by a factor of $1.3$, i.e., $16.4$ versus $12.38$. 
    
    \item \underline{\textit{The network disk volume usage}}: As Figure \ref{fig::network-disk-volume-B-vs-S5} shows, \starlit and the baseline demonstrate similar performance under this metric, i.e., $5.1$ compared to $4.98$. 
    
    \item \underline{\textit{The network file volume usage}}: As Figure \ref{fig::network-file-volume-B-vs-S5} demonstrates, under this metric, \starlit underperform the baseline by a factor of $1.3$, i.e., $1316$ versus $993$. 
    
\end{itemize}

 Hence, when the tree's depths in \starlit and baseline are set to  $5$ and $3$ respectively, then  \starlit can attain a superior AUPRC level compared to the baseline. However, in this setting, \starlit would impose approximately $1.3$ times higher cost than the baseline does.

% !TEX root =Feather.tex

%\vspace{-1mm}
\section{Further Applications of Starlit}\label{sec::other-usecases}
%\todo{this section needs an update... instead of string distance and geographical distance, which we didn't build, maybe we can talk about the game framework and the software pieces we built that can generalise...}

%Although the protocols and improvements proposed above are developed with the financial crime use case in mind, we expect the work to generalise to a broad range of settings. %For example, advances in secure string distance computations and geographical distance computation can be easily adapted to other applications that involve location data and/or free text information.% 
%Implementing SecureBoost with differential privacy and the other privacy protection measures we proposed in this paper can be helpful in advancing the current state of the art for learning gradient boosted decision trees when data is vertically partitioned across organisations.

%The solution described in this paper is highly adaptable to many use cases. The financial use case showcased here involves both the bank account data that is horizontally partitioned across a number of banks and \fsp transaction data representing data about the same individuals, but vertically partitioned across \fsp and the banks. 

\starlit demonstrates notable adaptability across a diverse array of collaborative analysis tasks, as outlined in the subsequent sections. Additionally, \starlit's components may have other applications, as discussed in Appendix \ref{sec::components-applications}.

% \item \underline{Mitigating Terrorism.}

%%\vspace{-2mm}
\subsection{Mitigating Terrorism}

The effective response to the challenge of counter-terrorism calls for the establishment of collaborative partnerships among diverse crime agencies at both national and international levels. 
While various countries may occasionally exchange some of their intelligence in plaintext, factors such as mishandling or leakage of (top) secret data \cite{top-secret-leakage-23,Trump-Biden-Clinton-data-leakage} can dissuade them from doing so. 

The facilitation of such collaboration in a \textit{privacy-preserving} manner can be achievable through the utilization of \starlit. By leveraging \starlit, crime agencies can engage in joint efforts, sharing their knowledge to create a unified intelligence model \cite{royalsocpets}. This collaborative approach involves the integration of data from different crime agencies, allowing for a more comprehensive understanding of potential threats. Specifically, each crime agency can augment its own dataset with flags assigned by counterpart agencies to individuals deemed suspicious, whether citizens or tourists. These flags are based on factors such as a person's history of violence or associations with known terrorist organizations. 

In this framework, \starlit can also empower crime agencies to identify disparities in the information provided by certain individuals across different countries' crime agencies. The exchange of information facilitated by \starlit in real-time contributes to a more comprehensive and accurate understanding of individuals with potential security implications. %This, in turn, strengthens the collective security efforts of the international community.

\subsection{Enhancing Digital Health}
While digital healthcare offers numerous advantages, addressing one of its primary challenges, namely data privacy, is crucial \cite{filkins2016privacy}. 
\starlit holds the promise of delivering advantages to the digital health sector. For example, it can empower hospitals, local medical practitioners, and wearable devices that gather diverse patient data. 
This data may encompass crucial indicators/flags revealing a patient's susceptibility to chronic diseases or their ongoing management of such conditions \cite{HosseiniHQHT23}. \starlit enables the identification of common patients among each of these entities and the active party responsible for developing a global model. They can then construct a model using the shared data and flags.

Within this framework, \starlit is capable of pinpointing disparities in features attributed to overlapping patients among different entities. For example, it can ascertain whether each involved party prescribed distinct medications for the same patient, afflicted with the same disease, at varying points in time. This unique capability plays a pivotal role in the detection of medication errors and streamlines the process of medication reconciliation \cite{Medication-Reconciliation}, all while upholding the vital principle of privacy.

%\vspace{-2mm}
\subsection{Detecting Benefit Fraud}

This process aims to empower different entities, such as government agencies, social service organizations, and different countries' banks to collaboratively develop a model to deal with benefit fraud \cite{Benefit-fraud}, which is against the law in certain countries.  
In this particular scenario, \starlit serves as the enabling technology, facilitating a government organization to develop a collaborative model with both national and international banks. Each participating bank can assign a distinctive flag to each customer's account, indicating whether the account is receiving social benefits from the respective country or is deemed suspicious.

% \starlit will allow a governmental organisation to develop a model in collaboration with various national and international banks, each of which may allocate a flag to each customer's account that indicates whether it receives social benefits from that country or is deemed suspicious. 
%

Much like its original application, \starlit excels in identifying disparities in the information provided by a customer to different entities. In this context, technology plays a crucial role in preventing fraud or misuse, ensuring that resources are allocated with fairness and appropriateness as top priorities.

% In this case (similar to the original application of \starlit), \starlit could identify discrepancies among the information that a customer provides to different parties. This approach will ultimately prevent fraud or misuse, and ensure that resources are allocated fairly and appropriately.

% \begin{itemize}[label=$\bullet$,leftmargin=4.5mm]

% \item \underline{Improving Malware Identification.} 
% Via collaboration among independent antivirus companies, each of which holds a proprietary database of malware, it is possible to improve malware detection while preserving the privacy of each company's databases. Traditionally, antivirus companies operate in isolation, relying solely on their internal databases to detect and combat malware threats. However, by coming together using a privacy-preserving collaborative mechanism, such as \starlit, these companies can share their unique datasets, providing a more comprehensive and updated understanding of the evolving landscape of malware.

% \end{itemize}

%Our solution allows the privacy-preserving computation of query results and the creation of features that originate from any one of the contributing parties (e.g. the average transaction amount for the sender in a given currency coming from \fsp data only, and the account flag coming from bank data only). 

%\vspace{-1mm}
\section{Conclusion and Future Work}\label{sec:Future-Work}
In this work, we introduced \starlit, a scalable privacy-preserving and demonstrated its applications in dealing with financial fraud, mitigating terrorism, and improving digital health. 
We formally defined and proved the security of \starlit in the simulation-based model. To formally capture the security of \starlit, we have defined a set of leakage functions that may hold independent significance.  
We implemented \starlit and conducted a comprehensive analysis of its performance and accuracy, using synthetic data provided by one of the key players facilitating financial transactions worldwide. 

%During the implementation of \starlit we had to overcome several challenges to build the implementation upon existing open-source FL frameworks, i.e., FATE and \flower. In this work, we reported these challenges and the techniques we used to overcome them, with the hope that they can assist future researchers who need to develop similar systems. 

In any secure FL, the output inevitably discloses certain information about participating parties' private inputs. This fact may dissuade some parties with sensitive and valuable inputs from engaging in the FL process, particularly when they lack interest in the outcome. Future research could enhance \starlit by rewarding active contributors, which could also bridge the gap between the data market \cite{golob2023decentralized,KochKMMR22,abs-2210-08723} and FL. Another avenue is strengthening \starlit's security against fully malicious parties.

\bibliographystyle{ACM-Reference-Format}
\bibliography{ref}

\appendix

% !TEX root =main.tex

\section{Related Work}\label{sec::survey}

In this section, we discuss the approaches used to deal with fraudulent financial transactions. This includes:

\begin{itemize}[leftmargin=4.2mm,label=$\bullet$]
    \item A centralized approach, where a client (e.g., bank) either trains its model based on the history of financial transactions it holds, or a centralized party (server) collects data in plaintext from different sources without considering the privacy of different parties. This approach is covered in Section \ref{sec:Dentralized-Approaches}.

    \item A decentralized approach that involves multiple entities, such as different banks and financial service providers (e.g., Visa or SWIFT), where parties collaboratively train their models while preserving the privacy of input data. This approach is discussed in Section \ref{sec:sec:Distributed-Approaches}.

% \item VFL-based Anomaly Detection Solutions, discussed in Section \ref{sec::VFL-based-Anomaly-Detection}.

    \item Other none AI-based schemes, which is covered in Section \ref{sec:sec:Beyond-AI}.

\end{itemize}

\subsection{Centralized Approaches Without Privacy Support}\label{sec:Dentralized-Approaches}

Afriyie  \textit{et al.} \cite{afriyie2023supervised} studied the performance of three different machine learning models, logistic regression, random forest, and decision trees to classify, predict, and detect fraudulent credit card transactions. They conclude that random forest produces maximum accuracy in predicting and detecting fraudulent credit card transactions. Askari and Hussain \cite{AskariH20} aim to achieve the same goal as Afriyie \textit{et al.} (i.e., to classify, predict, and detect fraudulent credit card transactions) using ``Fuzzy-ID3'' (Interactive Dichotimizer 3). Saheed \textit{et al.} \cite{saheed2022big} examined machine learning models for predicting credit card fraud and proposed a new model for credit card fraud detection by relying on principal component analysis and supervised machine learning techniques such as K-nearest neighbor, ridge classifier, and gradient boost. 

Srokosz \textit{et al.} \cite{srokosz2023machine} designed a mechanism to improve the rules-based fraud prevention systems of banks. The proposed mechanism is a rating system that uses unsupervised machine learning and provides early warnings against financial fraud. The proposed system basically distinguishes between rogue and legitimate bank account login attempts by examining customer logins from the banking transaction system. The suggested method enhances the organization’s rule-based fraud prevention system. Al-Abri \textit{et al.} \cite{al2021improving} proposed a data mining mechanism based on logistic regression to detect irregular transactions, implemented the solution, and analyzed its performance. We refer readers to \cite{cao2022ai} for a survey of AI-based mechanisms used in traditional financial institutions.

Researchers have also focused on financial reports/statements of companies and developed financial statement fraud detection mechanisms for Chinese listed companies using deep learning, e.g., in \cite{wu4292653optimized,xiuguo2022analysis}. Their threat model and solution considered companies as misbehaving actors who want to convince investors, auditors, or governments that they have followed the regulations and that their financial statements are valid. 

Researchers also proposed data mining-based mechanisms to provide a detection model for Ponzi schemes on the Ethereum blockchain, e.g., see \cite{jung2019data,chen2019exploiting}. Very recently, Aziz \textit{et al.} \cite{aziz2023modified}  proposed an optimization strategy for deep learning classifiers to identify fraudulent Ethereum transactions and Ponzi schemes. We refer readers to \cite{hassan2022anomaly} for a survey of anomaly detection in the blockchain network. Note that the proposed solutions for blockchain cannot be directly applied to the conventional banking system as it is in a different setting.

\subsection{Distributed Approaches With Privacy Support}\label{sec:sec:Distributed-Approaches}

% Federated Learning (FL) is a machine learning paradigm where multiple parties collaboratively build machine learning models without directly revealing their sensitive input data to their counterparts \cite{YangLCT19}. The concept of FL was initially proposed by Google in 2016 \cite{McMahanMRA16} which primarily focused on a scenario where millions of smartphones are coordinated by a central server while local data of the smartphones are not transmitted to any party. 
% %
Generally, FL can be categorized into three classes \cite{YangLCT19}, according to how data is partitioned:
\begin{itemize}[label=$\bullet$]
    \item Horizontal Federated Learning (HFL).
    \item Vertical Federated Learning (VFL).
    \item Federated Transfer Learning (FTL).
\end{itemize}

HFL refers to the FL setting where participants share the same feature space while
holding different samples.  On the other hand, VFL refers to the FL setting where datasets share the same samples while holding different features. FTL refers to the FL setting where datasets differ in both feature and sample spaces
with limited overlaps. 
For the remainder of this section, our focus will mainly be on FL-based schemes developed to deal with fraudulent transactions.

\subsubsection{HFL-Based Approaches.} 

Suzumura \textit{et al.} \cite{abs-1909-12946} developed an ML-based privacy-preserving mechanism to share key information across different financial institutions. This solution builds 
ML models by leveraging FL and graph learning. This would ultimately allow for global financial crime detection while preserving the privacy of financial organizations and their customers' data. Given that federated graph learning involves collaborative training on a shared graph structure (common set of features) distributed across multiple parties, this proposed solution aligns with the characteristics of HFL.  
Moreover, Yang \textit{et al.} \cite{yang2019ffd} proposed an FL-based method using ``Federated Averaging'' \cite{McMahanMRHA17} to train a model that can detect credit card fraud. Similar to the scheme in \cite{abs-1909-12946}, this method falls under the HFL category. 

Unlike the schemes explained above, \starlit deals with the data that have been partitioned both vertically and horizontally.

%%%%% check if there're more HFL-based schemes

 \subsubsection{VFL-Based Approaches.}

Simultaneously with our solution, another approach has been developed by Qiu \et. \cite{abs-2305-11236}. This alternative relies on neural networks and aims to deal with financial fraud. It strives for computational efficiency primarily through the use of symmetric key primitives. The scheme incorporates the elliptic-curve Diffie-Hellman key exchange and one-time pads to secure exchanged messages during the model training phase. This solution has also been implemented and subjected to performance evaluation.

\noindent\underline{\textit{\starlit versus the Scheme of Qiu \et.}}  The latter scheme requires each client (e.g., bank) to possess knowledge of the public key of every other client and compute a secret key for each through the elliptic-curve Diffie-Hellman key exchange scheme. Consequently, this approach imposes $O(n)$ modular exponentiation on each client, resulting in the protocol having a complexity of  $O(n^{\st 2})$, where $n$ represents the total number of clients. In contrast, in \starlit, each client's complexity is independent of the total number of clients and each client does not need to know any information about other participating clients.  
Moreover, the scheme proposed in \cite{abs-2305-11236} assumes the parties have already performed the identity alignment phase, therefore, the implementation, performance evaluation, and security analysis exclude the identity alignment phase.

Furthermore, the scheme in \cite{abs-2305-11236} fails to terminate successfully even if only one of the clients neglects to transmit its message. In this scheme, each client, utilizing the agreed-upon key with every other client, masks its outgoing message with a vector of pseudorandom blinding factors. The expectation is that the remaining clients will mask their outgoing messages with the additive inverses of these blinding factors. These blinding factors are generated such that, when all outgoing messages are aggregated, the blinding factors cancel each other out. Nevertheless, if one client fails to send its masked message, the aggregated messages of the other clients will still contain blinding factors, hindering the training on correct inputs. 
The solution proposed in \cite{BonawitzIKMMPRS17}, based on threshold secret sharing, can address this issue. However, incorporating such a patch would introduce additional computation and communication overheads. In contrast, \starlit does not encounter this limitation. This is because the message sent by each client is independent of the messages transmitted by the other clients.

Lv \et. \cite{LvCZYMW21} introduced a VFL-based approach to identify black market fraud accounts before fraudulent transactions occur. This approach aims to guarantee the safety of funds when users transfer funds to black market accounts, enabling the financial industry to utilize multi-party data more efficiently.  
The scheme involves data provided by financial and social enterprises, encompassing financial features extracted from a bank such as mobile banking login logs, and account transaction information. Social features include the active cycle corresponding to the mobile phone number, the count of malicious apps, and the frequency of visits to malicious sites. The approach utilizes \textit{insecure} hash-based PSI for identity alignment.

This scheme differs from \starlit in a couple of ways:  (i) \starlit operates in a multi-party setting, where various clients contribute their data, in contrast to the aforementioned scheme, which involves only two parties and cannot be directly applied to the multi-client setting, and (ii) \starlit deals with the data that has been partitioned both horizontally and vertically, whereas the above scheme focuses only on vertically partitioned data.

\subsubsection{FL-Based Solutions to Detect Fraudulent Financial Transactions when Datasets are Vertical and Horizontally Partitioned.}

% Lv \et \cite{LvCZYMW21} proposed a VFL-based approach to identify black market fraud accounts before fraudulent transactions occurs and protect the safety of funds when users transfer funds to black market fraud accounts.  This would enable the financial industry to use multi-party data more efficiently for anti-fraud research. 

Recently, to combat the global challenge of organized crime, such as money laundering, terrorist financing, and human trafficking,  the UK and US governments launched a set of prize challenges \cite{PETs2023}. This competition encouraged innovators to develop technical solutions to identify suspicious bank account holders while preserving the privacy of honest account holders by relying on FML and cryptography approaches. This underscores the importance the distributed privacy-preserving financial data analytics for governments.

Very recently, in parallel with our work, Arora \et.  \cite{abs-2310-04546} introduced an FL-based approach for detecting anomalous financial transactions. This methodology was specifically devised for the aforementioned prize challenge and relied on oblivious transfer, secret sharing, differential privacy, and multi-layer perception. The authors have implemented the solution and conducted a thorough analysis of its performance and accuracy.

\noindent\underline{\textit{\starlit versus the Scheme of Arora \et.}}
The latter assumes that the ordering bank never allows a customer with a dubious account to initiate any transactions, but it permits the same account to receive money. In simpler terms, this scheme exclusively deals with frozen accounts, limiting its applicability. 
This approach will exempt the
ordering bank from participating in MPC, which ultimately results in lower overhead (compared to the case where the ordering bank had to be involved in MPC). 

In the real world, users' accounts might be deemed suspicious (though not frozen), yet they can still conduct financial transactions within their bank. However, the bank may handle such accounts more cautiously than other non-suspicious accounts.  In contrast, in the context of dealing with financial transactions, \starlit does not place any assumption on how a bank treats a suspicious account.

Furthermore, unlike the scheme proposed in \cite{abs-2310-04546}, which depends on an ad-hoc approach to preserve data privacy during training, \starlit, employs SecureBoost, which is a well-known scheme extensively utilized and analyzed in the literature. 
Thus, compared to the scheme in \cite{abs-2310-04546}, \starlit considers a more generic scenario and relies on a more established scheme for VFL.

Kadhe \et. \cite{abs-2310-19304} proposed a privacy-preserving anomaly detection scheme, that relies on fully homomorphic encryption (highly computationally expensive), DP, hash tables, and secure multi-party computation. Similarly, the authors have implemented their solution and analyzed its performance.

\noindent\underline{\textit{\starlit versus the Scheme of Kadhe \et.}} The scheme in \cite{abs-2310-19304} heavily relies on fully homomorphic encryption. In this setting, all parties need to perform fully homomorphic operations, by performing either computation on the outputs of the homomorphic encryption or encrypting and decrypting ciphertexts. This will affect both the scalability and efficiency of this scheme. In contrast, \starlit refrains from using any \textit{fully} homomorphic scheme. 

%Furthermore, the security of this scheme is only informally proved without even defining the security of the scheme in the first place. In contrast, \starlit does not use any fully homomorphic encryption, also it is accompanied by formal security definition and proof.   

All of the above solutions share another shortcoming, they lack formal security definitions and proofs of the proposed systems.

%%% talk about the paper by Cambridge
%%% there might be a paper by WeBank-- check it 

%\subsubsection{VFL-Based Solution to Deal with Financial Crime.}
\subsubsection{VFL-Based Solutions to Detect Anomaly in Other Contexts}\label{sec::VFL-based-Anomaly-Detection}

We \et. \cite{abs-2205-14196} proposed a privacy-preserving solution specifically designed to detect anomalies within multiple attributed networks in a privacy-preserving manner, e.g., to detect abnormal IPs or predict cyber attacks. The authors have evaluated the proposed solution using various real datasets, including Computer networks, Car-hailing, Bicycle-sharing, Subway traffic, and Point-of-Interest datasets.
However, this solution lacks generality and has been designed specifically for a very specific setting.

\subsection{Solutions Beyond AI}\label{sec:sec:Beyond-AI}
There have also been efforts to deal with banking fraud, by using alternative (none AI-based) prevention mechanisms, such as Multi-Factor Authentication (M-FA) and Confirmation of Payee (CoP) schemes, outlined below. 

For bank users to prove their identity to remote service providers or banks, they provide a piece of evidence, called an ``authentication factor''. Authentication factors can be based on (i) knowledge factors, e.g., PIN or password, (ii) possession factors, e.g., access card or physical hardware token, or (iii) inherent factors, e.g., fingerprint. Knowledge factors are still the most predominant factors used for authentication \cite{bonneau2010password,murdoch2022forward}. The knowledge factors themselves are not strong enough to adequately prevent impersonation \cite{jacomme2021extensive,sinigaglia2020survey}. M-FA methods that depend on more than one factor are more difficult to compromise than single-factor methods. Thus, in general, M-FA methods lower the chance of fraudsters who want to gain unauthorized access to users’ online banking accounts. 

Confirmation of Payee (CoP) is a name-checking service for UK-based payments \cite{CoP,AbadiM23}. It provides customers greater assurance that they are sending payments to the intended recipient, helping to avoid making accidental, misdirected payments to the wrong account holder, as well as providing another layer of protection in the fight against fraud, especially Authorised Push Payment fraud \cite{FCA-Glossary}. In short, CoP ensures that a money recipient's details (inserted by the money sender) match the record held by the recipient's bank. %Even though M-FA and CoP are effective mechanisms, they focus on very specific types of attack/misbehaviour and lack generality. 

\section{Proof of Security}\label{sec::Proof-of-Security}

\begin{proof} We prove Theorem \ref{theo::starlit-privacy} by examining the scenario in which each party is corrupt, one at a time.

\subsection{Corrupt \aut}\label{sec::Corrupt-FSP} 
%\textbf{Corrupt \fsp.} 

In the real execution, the view of \aut is defined as follows: 

%{\small{
$$\mathsf{View}_{\st \aut}^{\st \adv, \starlit}\Big(\underbrace{\prm_{\st \aut}, DS_{\st \aut}}_{\st \text{Inputs of } \aut}, \underbrace{DS_{\st \bank_{\st 1}},..., DS_{\st\bank_{\st n}}}_{\st \text{Inputs of $\bank_{\st 1},...,\bank_{\st n}$}}, \underbrace{\prm_{\st\fc}}_{\st\text{Input of }{\fc}}\Big):=$$

$$\{\mathsf{View}_{\st \aut}^{\st \adv, \mathcal{PSI}_{\st 1}}, ...,  
 \mathsf{View}_{\st \aut}^{\st \adv, \mathcal{PSI}_{\st n}}, \mathsf{View}_{\st \aut}^{\st \adv, \mathcal{SB}} \}$$
%}}

 where $\prm_{\st \aut}$ includes the input parameters (e.g., the initial global model) of \aut to federated learning, $DS_{\st \aut}$ is the dataset held by \aut, $DS_{\st \bank_{\st i}}$ is a dataset held by $\bank_{\st i}$, and $\prm_{\st \fc}$ includes the inputs parameters of \fc to federated learning.

Furthermore, each $\mathsf{View}_{\st \aut}^{\st \adv, \mathcal{PSI}_{i}}$ refers to the \aut's real-model view during the execution of an instance of  $\mathcal{PSI}$ with $\bank_{\st i}$, while $\mathsf{View}_{\st \aut}^{\st \adv, \mathcal{SB}}$ is \aut's real-model view during the execution of SecureBoost. The simulator $\mathsf{Sim}^{\st \func,  \leak_{\st 1}}_{_{\st \aut}}$, which receives $\aut$'s inputs $(\prm_{\st \aut}$, $DS_{\st \aut})$ and the leakage $\leak_{\st 1}(inp):=\Big(v_{\st 1}, ..., v_{\st n},\leakW_{\st 1}(\prm_{\st 1},$ $ \prm_{\st 2})\Big)$ operates as follows. 

\begin{enumerate}[leftmargin=4.8mm]
    \item generates an empty view. 
    
    \item generates $n$ sets $DS'_{\st\bank_{\st 1}},..., DS_{\st\bank_{\st n}}'$, where the size of each $DS'_{\st\bank_{\st i}}$ is $v_{\st i}$ and the element of  $DS'_{\st\bank_{\st i}}$ are picked uniformly at random from the set elements' universe (for all $i$, $1\leq i\leq n$). 
    \item for each $\bank_{\st i}$, extracts the \aut-side simulation of $\mathcal{PSI}$ from the simulator of $\mathcal{PSI}$ using input sets $DS_{\st \aut}$ and $DS'_{\st\bank_{\st i}}$. Let $\mathsf{Sim}^{\st \mathcal{PSI}_{\st i}}_{_{\st \aut}}$ denote this simulation. Note that the latter simulation is guaranteed to exist because this specific PSI, i.e., $\mathcal{PSI}$, has been proven secure in \cite{KolesnikovKRT16}. It appends each $\mathsf{Sim}^{\st \mathcal{PSI}_{\st i}}_{_{\st \aut}}$ to the view. 
    \item extracts the \aut-side simulation of SecureBoost $\mathcal{SB}$ from the simulator of $\mathcal{SB}$. Let $\mathsf{Sim}^{\st \mathcal{SB}}_{_{\st \aut}}$ denote this simulation. It appends $\mathsf{Sim}^{\st \mathcal{SB}}_{_{\st \aut}}$ to the view. Note that we assume the latter simulation exists and can be produced, using the leakage $\leakW_{\st 1}(\prm_{\st 1}, \prm_{\st 2})$ and $\mathcal{SB}$ introduced in \cite{cheng2021secureboost}. It outputs the view.
\end{enumerate}

Now, we are prepared to demonstrate that the two views are indistinguishable. Since  $\mathcal{PSI}$ has been proven secure, $\mathsf{View}_{\st \aut}^{\st \adv, \mathcal{PSI}_{\st i}}$  and $\mathsf{Sim}^{\st \mathcal{PSI}_{\st i}}_{_{\st \aut}}$
 are computationally indistinguishable. Similarly, under the assumption that SecureBoost is simulatable,  $\mathsf{View}_{\st \aut}^{\st \adv, \mathcal{SB}}$ and $\mathsf{Sim}^{\st \mathcal{SB}}_{_{\st \aut}}$ are  distinguishable. Thus, the real and ideal models are indistinguishable.

\subsection{Corrupt \fc}\label{sec::Corrupt-FC} 

In the real execution, the view of \fc is defined as follows:
%
%\begin{equation*}
%\begin{split}
 $$\mathsf{View}_{\st \fc}^{\st \adv, \starlit}\Big({\prm_{\st \aut}, DS_{\st \aut}}, DS_{\st \bank_{\st 1}},..., DS_{\st\bank_{\st n}}, \prm_{\st \fc}\Big):=$$
 $$\{ S_{\st \bank_{\st 1}}, ..., S_{\st \bank_{\st n}}, \mathsf{View}_{\st \fc}^{\st \adv, \mathcal{SB}} \}  $$
 %\end{split}
%\end{equation*}

where $S_{\st \bank_{\st i}}$ is a set of size $s_{\st i}$. It contains triples each of which has the form $(ID_{\st \susr}, b_{\st \susr}, w_{\st \susr})$ corresponding to a user $\usr$, where $ID_{\st \susr}$ represents a random ID of a sample, $b_{\st \susr}$ is a differentially-private binary flag indicating features' inconsistency, $w_{\st \susr}$ is another differentially-private binary flag (indicating a suspicious user). The simulator $\mathsf{Sim}^{\st \func,  \leak_{\st 2}}_{_{\st \fc}}$, which receives leakage $\leak_{\st 2}(inp):=\Big(s_{\st 1},..., s_{\st n}, $ $\leakW_{\st 2}(\prm_{\st 1}, $ $\prm_{\st 2})\Big)$ operates as follows.

\begin{enumerate}[leftmargin=4.5mm]
    \item generates an empty view. 
    \item constructs $n$ sets $S'_{\st \bank_{\st 1}},..., S'_{\st \bank_{\st n}}$, where each $S'_{\st \bank_{\st i}}$ has the following form. 
    \begin{itemize}[label=$\bullet$]
        \item contains $s_{\st i}$ triples $(ID', b', w')$.
        \item each $ID'$ is a string picked uniformly at random. 
        \item each $b'$ has been constructed by picking a random binary value and then applying the local differential privacy to it. 
        \item each $w'$ is generated the same way as $b'$ is generated. 
    \end{itemize}
    \item appends $S'_{\st \bank_{\st 1}},..., S'_{\st \bank_{\st n}}$ to the view. 

    \item extracts the \fc-side simulation of $\mathcal{SB}$ from the simulator of $\mathcal{SB}$. Let $\mathsf{Sim}^{\st \mathcal{SB}}_{_{\st \fc}}$ denote this simulation. It appends $\mathsf{Sim}^{\st \mathcal{SB}}_{_{\st \fc}}$ to the view. We assume the latter simulation can be produced, using the leakage $\leakW_{\st 2}(\prm_{\st 1}, \prm_{\st 2})$ and $\mathcal{SB}$. It outputs the view.
\end{enumerate}

Next, we argue that the two views are indistinguishable. In the real model, each $ID_{\st \susr}$ is a uniformly random string, as is each $ID'$ in the ideal model. Thus, they are indistinguishable. Also, the elements of each pair $(b_{\st \susr}, w_{\st \susr})$ in the real model are the output of the differential privacy mechanism (DP), so are the elements of each pair $(b', w')$. Due to the security of DP, $(b_{\st \susr}, w_{\st \susr})$ and $(b', w')$ are differentially private and indistinguishable. Moreover, under the assumption that SecureBoost is simulatable,  $\mathsf{View}_{\st \fc}^{\st \adv, \mathcal{SB}}$ and $\mathsf{Sim}^{\st \mathcal{SB}}_{_{\st \fc}}$ are distinguishable.  Hence, the real and ideal models are indistinguishable. 

\subsection{Corrupt Client}\label{sec::Corrupt-bank} 

In the real execution, the view of each $\bank_{\st i}$ is defined as follows:

$$\mathsf{View}_{\st \bank_{\st i}}^{\st \adv, \starlit}\Big(({\prm_{\st \aut}, DS_{\st \aut})}, DS_{\st \bank_{\st 1}},..., DS_{\st\bank_{\st n}}, \prm_{\st \fc}\Big):=$$

$$\{(DS_{\st \aut}\ \cap\ DS_{\st \bank_{\st i}}), S_{\st \aut}, \mathsf{View}_{\st \bank_{\st i}}^{\st \adv,  \mathcal{PSI}} \}$$

The simulator $\mathsf{Sim}^{\st \func,  \leak_{\st i+2}}_{_{\st \bank_{\st i}}}$, which receives input $DS_{\st \bank_{\st i}}$ and leakage $\leak_{\st \st i+2}(inp):=\Big((DS_{\st \aut}\ \cap\ DS_{\st \bank_{\st i}}), |DS_{\st \aut}|, S_{\st \aut} \Big)$ operates as follows.

\begin{enumerate}
    \item generates an empty view and appends $DS_{\st \aut}\ \cap\ DS_{\st \bank_{\st i}}$ to the view. 

    \item\label{bank-side-sim-step-gen-pair}  sets $z=|{S}_{\st \aut}|$. Then, it generates $z$ pairs, each of which has the form $(ID', \feat'_{\st \susr})$. In each pair, $ID'$ is a uniformly random string (picked from the IDs' universe), and $\feat'_{\st \susr}$ is one of the user's unique features in $S_{\st \aut}$. 
    
    \item generates an empty set $\bar{S}_{\st \aut}$ and appends all the above $z$ pairs (generated in step \ref{bank-side-sim-step-gen-pair}) to $\bar{S}_{\st \aut}$. It appends $\bar{S}_{\st \aut}$ to the view. 

    \item generates an empty set $\bar{DS}_{\st \aut}$ and then appends $DS_{\st \aut}\ \cap\ DS_{\st \bank_{\st i}}$ to $\bar{DS}_{\st \aut}$. 
    
    \item appends to  $\bar{DS}_{\st \aut}$ a set of uniformly random elements (from the set elements' universe) such that the size of the resulting set $\bar{DS}_{\st \aut}$ is $|DS_{\st \aut}|$. 
    
    \item for each $\bank_{\st i}$, extracts the $\bank_{\st i}$-side simulation of $\mathcal{PSI}$ from the simulator of $\mathcal{PSI}$ using input sets $\bar{DS}_{\st \aut}$ and $DS_{\st\bank_{\st i}}$. Let $\mathsf{Sim}^{\st \mathcal{PSI}_{\st i}}_{_{\st \aut}}$ denote this simulation. It appends each $\mathsf{Sim}^{\st \mathcal{PSI}_{\st i}}_{_{\st \aut}}$ to the view. It outputs the view. 

\end{enumerate}

Now, we are ready to demonstrate that the two views are indistinguishable. The intersection $DS_{\st \aut}\ \cap\ DS_{\st \bank_{\st i}}$ is identical in both views; therefore, they have identical distributions. Each $ID$ in $S_{\st \aut}$ is a uniformly random string, so is each $ID'$ in $\bar{S}_{\st \aut}$. As a result, they have identical distributions too. Also, each user's unique feature $\feat_{\st c}$ in $S_{\st \aut}$ and  $\feat'_{\st c}$ in $\bar{S}_{\st \aut}$ have identical distributions, as they are both picked from  ${S}_{\st \aut}$ that includes some users' of $\bank_{\st i}$. Due to the security of  $\mathcal{PSI}$, $\mathsf{View}_{\st \aut}^{\st \adv, \mathcal{PSI}_{\st i}}$  and $\mathsf{Sim}^{\st \mathcal{PSI}_{\st i}}_{_{\st \aut}}$
 are computationally indistinguishable. We can conclude that the two views are computationally indistinguishable.

\end{proof}

\section{Further Discussion on PSI Implementation}\label{sec:Further-Discussion-PSI-Implementation}

In this work, we initially focused on and implemented the RSA-based PSI proposed in \cite{agrawal2003information}, due to its simplicity that would lead to easier security analysis. However, after conducting a concrete run-time analysis we noticed that this protocol has a very high run time.  Therefore, we adjusted and used the Python-based implementation of the PSI introduced in \cite{KolesnikovKRT16} which yields a much lower run-time than the one in \cite{agrawal2003information}. This protocol mainly relies on efficient symmetric key primitives, such as Oblivious Pseudorandom Functions and  Cuckoo hashing. The security of this PSI has been proven in a standard simulation-based (semi-honest) model and the related paper has been published at a top-tier conference. The efficiency of the PSI protocol mainly stems from the efficient ``Oblivious Pseudorandom Function'' (OPRF) that Kolesnikov \textit{et al.} constructed which itself uses the Oblivious Transfer (OT) extension proposed earlier in \cite{KolesnikovK13}.

To compare the two PSIs' runtime we conducted certain experiments, before developing the \starlit's implementation. The experiments were carried out on a laptop with 8 cores, 2.4 GHZ i9 CPU and 64GB of memory. We did not take advantage of parallelization.
Our experiments were carried out with each party having $2^{\st n}$ set elements and compared the run-time of the PSI in \cite{agrawal2003information} with the one in \cite{KolesnikovKRT16}. We concluded that the PSI in \cite{KolesnikovKRT16} is around 10--11 times faster than the one in \cite{agrawal2003information}. 
We conducted the experiments when each party's set's cardinality is in the range $[2^{\st 9},2^{\st 19}]$. Briefly, our evaluation indicates that the PSI's runtime increases from 0.84 to 367.93 seconds when the number of elements increases from $2^{\st 9}$ to $2^{\st 19}$. 
Table \ref{table::run-time-comparison} summarizes the run-time comparison between the two PSIs.  We use the Flower adapter to communicate between the PSI client (i.e., \fsp) and the PSI server (i.e., a bank \bank). An instance of the PSI, for each account, takes as input the following string: 
$$account_{\st number} || customer_{\st name} || street_{\st name} || countryCity_{\st zipcode}$$

% The output of the PSI is received by the participating \bank.

% !TEX root =Feather.tex

%\vspace{-4mm}
\begin{table}[!ht]
%\begin{footnotesize}
\begin{center}
\caption{\small The  run-time comparison between the RSA-based PSI in \cite{agrawal2003information} and our choice of PSI proposed in \cite{KolesnikovKRT16} (in sec.).}\label{table::run-time-comparison}

\renewcommand{\arraystretch}{1}

\scalebox{0.79}{
\begin{tabular}{|c|c|c|c|c|c|c|c|c|c|c|c|c} 
\hline 
%{\rowcolors{2}{blue!80!gray!18}{blue!10!gray!18}

\cellcolor{\clr}&\multicolumn{11}{c|}{\cellcolor{\clr}\scriptsize Set Cardinality}\\

\cline{2-12}

\cline{2-12}

\multirow{-2}{*} {\cellcolor{\clr}\scriptsize Protocol}&\cellcolor{white!20}\scriptsize$\scriptsize2^{\scriptscriptstyle 9}$ &\cellcolor{white!20}\scriptsize $2^{\scriptscriptstyle 10}$&\cellcolor{white!20}\scriptsize$2^{\scriptscriptstyle 11}$&\cellcolor{white!20}\scriptsize$2^{\scriptscriptstyle 12}$&\cellcolor{white!20}\scriptsize $2^{\scriptscriptstyle 13}$&\cellcolor{white!20}\scriptsize$2^{\scriptscriptstyle 14}$&\cellcolor{white!20}\scriptsize$2^{\scriptscriptstyle 15}$&\cellcolor{white!20}\scriptsize$2^{\scriptscriptstyle 16}$& \cellcolor{white!20}\scriptsize$2^{\scriptscriptstyle 17}$&\cellcolor{white!20}\scriptsize$2^{\scriptscriptstyle 18}$&\cellcolor{white!20}\scriptsize$2^{\scriptscriptstyle 19}$ \\
    \hline
    
    \hline

%\cite{eopsi}
\scriptsize\cite{agrawal2003information}&\cellcolor{gray!20} \scriptsize {4.10} &\cellcolor{gray!20} \scriptsize {9.32} &\cellcolor{gray!20}  \scriptsize {16.56} &\cellcolor{gray!20}\scriptsize {32.78} &\cellcolor{gray!20}\scriptsize   {65.45} &\cellcolor{gray!20}\scriptsize  {132.97} &\cellcolor{gray!20}\scriptsize  {252.56} &\cellcolor{gray!20}\scriptsize  {524.32}  &\cellcolor{gray!20}\scriptsize{1059.49} &\cellcolor{gray!20}\scriptsize {-} & \cellcolor{gray!20}\scriptsize {-}    \\

    \hline

\scriptsize\cite{KolesnikovKRT16}&\cellcolor{gray!20} \scriptsize {0.84}&\cellcolor{gray!20} \scriptsize {1.18} &\cellcolor{gray!20}  \scriptsize  {1.84} &\cellcolor{gray!20}\scriptsize {3.23} &\cellcolor{gray!20}\scriptsize   {6.06} &\cellcolor{gray!20}\scriptsize  {11.63} &\cellcolor{gray!20}\scriptsize   {23.75} &\cellcolor{gray!20}\scriptsize  {45.80}  &\cellcolor{gray!20}\scriptsize {99.09} &\cellcolor{gray!20}\scriptsize {183.79} & \cellcolor{gray!20}\scriptsize   {367.93}  \\

       \hline

\end{tabular}
}
\end{center}
%\end{footnotesize}
\end{table}

\section{Fields of Dataset 1}\label{secc:DB-1-fields}

The provided synthetic Dataset 1 contains the following fields:

\begin{itemize}[leftmargin=4.5mm,label=$\bullet$]

    \item \underline{MessageId:}  Globally unique identifier within this dataset for individual transactions.
    \item \underline{UETR:} The Unique End-to-end Transaction Reference—a 36-character string enabling traceability of all individual transactions associated with a single end-to-end transaction
    \item \underline{TransactionReference:}  Unique identifier for an individual transaction.

    \item \underline{Timestamp:} Time at which the individual transaction was initiated.

    \item \underline{Sender:} Institution (bank) initiating/sending the individual transaction.

    \item \underline{Receiver:} Institution (bank) receiving the individual transaction.

    \item \underline{OrderingAccount:} Account identifier for the originating ordering entity (individual or organization) for end-to-end transaction.

    \item \underline{OrderingName:} Name for the originating ordering entity.

    \item \underline{OrderingStreet:} Street address for the originating ordering entity.

    \item \underline{OrderingCountryCityZip:}  Remaining address details for the originating ordering entity.

    \item \underline{BeneficiaryAccount:} Account identifier for the final beneficiary entity (individual or organization) for end-to-end transaction.

    \item \underline{BeneficiaryName:} Name for the final beneficiary entity.

    \item \underline{BeneficiaryStreet:} Street address for the final beneficiary entity.

    \item \underline{BeneficiaryCountryCityZip:}  Remaining address details for the final beneficiary entity.

    \item \underline{SettlementDate:} Date the individual transaction was settled.

    \item \underline{SettlementCurrency:}  Currency used for transaction.

    \item \underline{SettlementAmount:} Value of the transaction net of fees/transfer charges/forex.

    \item \underline{InstructedCurrency:}  Currency of the individual transaction as instructed to be paid by the Sender.

    \item \underline{InstructedAmount} Value of the individual transaction as instructed to be paid by the Sender.

    \item \underline{Label:}  Boolean indicator of whether the transaction is anomalous or not. This is the target variable for the prediction task.

\end{itemize}

\section{Fields of Dataset 2}\label{secc:DB-2-fields}

The provided synthetic Dataset 2 contains the following fields:

\begin{itemize}[label=$\bullet$]

\item \underline{Bank:} Identifier for the bank.

\item \underline{Account:} Identifier for the account.

\item \underline{Name:} Name of the account.

\item \underline{Street:} Street address associated with the account.

\item \underline{CountryCityZip:}  Remaining address details associated with the account.

\item \underline{Flags:} Enumerated data type indicating potential issues or special features that have been associated with an account.  

\end{itemize}

\section{Challenges}\label{sec::challenges}

To implement \starlit, we encountered a set of challenges. In this section, we briefly discuss these challenges and explain how we overcame them, with the hope that they can assist future researchers who need to develop similar systems.

\subsection{The Challenge of Using \flower in Starlit's Implementation}

The \flower clashed with our proposed architecture in which there was no central server controlling the process. 
%
% Instead, clients directly communicate with one another to exchange messages. 
%
In \starlit, \fsp acts as a coordinator responsible for initiating the feature extraction and learning phases. 
Nevertheless, in \flower, a central server executes a series of training rounds, each comprising a sequence of steps that involve sending instructions to and receiving results from the clients in the system. \flower also demands the server's strategy to have precise knowledge of each client's designated actions in every round of communication.

This process complicates the ability to perform any pre-processing of data before entering the iterative training process, a necessity for us to complete our feature extraction phase.

To address these challenges, we developed a server strategy implementation that functions as a message router for the clients in the system. This design enabled us to preserve our initial concept of  \aut acting as a coordinating entity, allowing the server strategy to remain indifferent to the specific actions required in each round of messages. This alteration significantly streamlined the server logic. Moreover, this change substantially enhances the prototype's extensibility and adaptability to additional use cases.

\subsection{The Challenge of Using FATE in Starlit's Implementation}
The implementation of \starlit uses SecureBoost implementation initially integrated within FATE, an open-sourced FL project (as discussed in Section \ref{sec:SecureBoost}, page \pageref{sec:SecureBoost}). In \starlit, SecureBoost is executed jointly among \aut and \fc. 
Since \flower does not incorporate FATE and the \flower API itself does not permit parties to communicate directly with each other, we have devoted significant effort to readjusting the implementation of \starlit to seamlessly integrate with the \flower API. We address this challenge by routing the messages that parties exchange through the \flower's server.

Moreover, the \flower's clients operate in a stateless manner. To seamlessly integrate with FATE without the necessity to snapshot the entire system for each \flower round, we developed a new implementation of the FATE API. This implementation utilizes Python multiprocessing queues for sending and receiving messages. By adopting this approach, we were able to instantiate longer-lived standalone FATE processes dedicated to training and prediction tasks. In this configuration, each \flower client acts as a proxy, facilitating the exchange of messages with the respective FATE process through Python multiprocessing queues.%, as illustrated in Figure \ref{fig:fate-flower-comms}. %The end result each party can communicate with each other by routing the messages through the \flower server (as the original \flower implementation requires). 

% \begin{figure}
%     \centering
%     \includegraphics[width=0.9\textwidth]{images/fate_flower.png}
%     \caption{\small In order to integrate the FATE federated AI framework with the Flower framework used in the competition, our solution runs a separate FATE process for each of the parties involved in the vertically partitioned learning. The Flower Client communicates with the FATE process using python multiprocessing queues. All messages between the parties are routed via Flower.}
%     \label{fig:fate-flower-comms}
% \end{figure}

\section{Starlit's CPU Utilization}\label{sec::CPU-Utilization}

\begin{figure}[!h]
    \centering
    \includegraphics[width=0.35\textwidth]{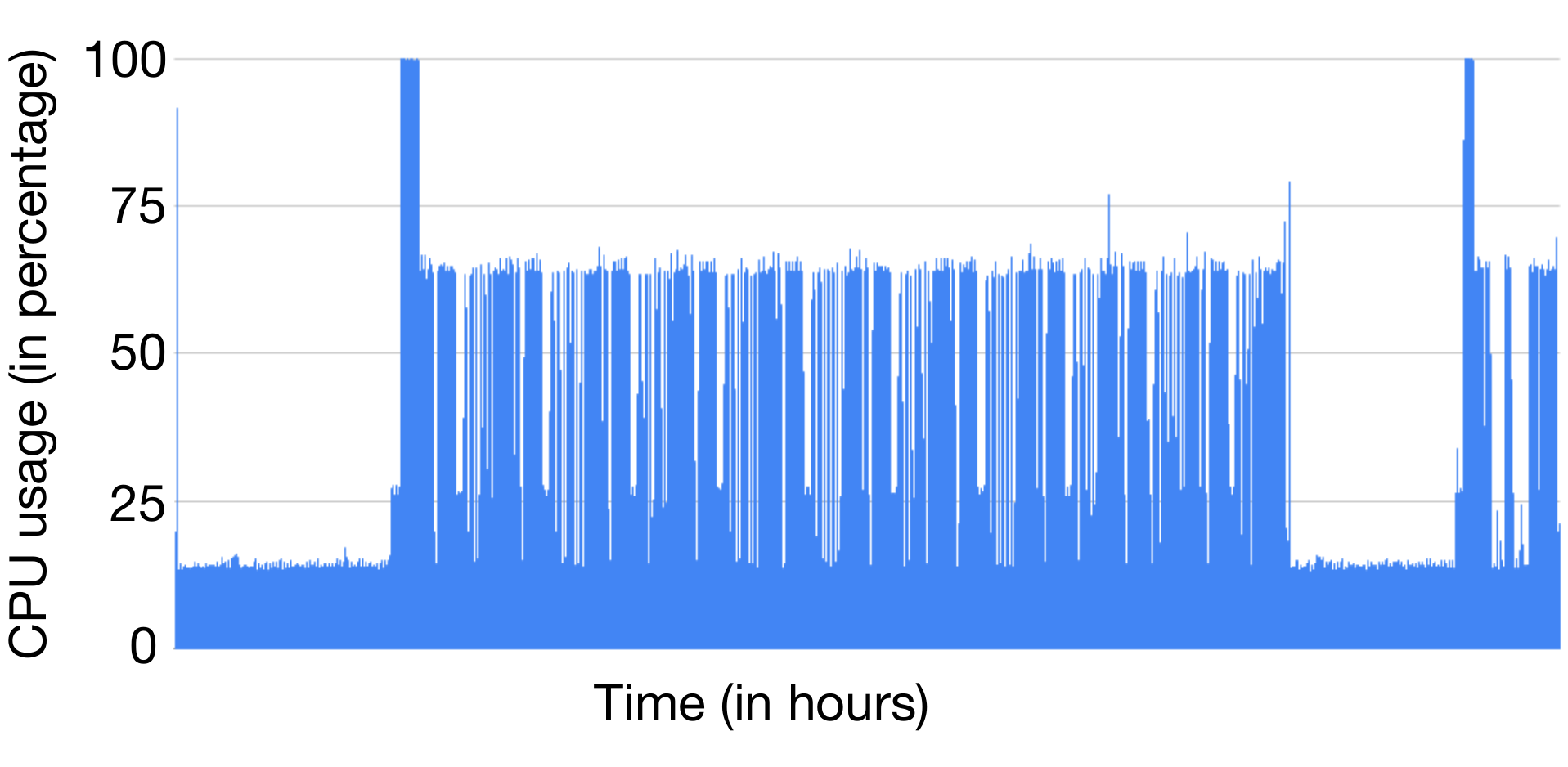}
    \caption{\small CPU utilisation of \starlit. Peaks in the graph indicate times of high CPU activity, while valleys represent periods of lower activity. A steady line with little variation indicates a consistent level of CPU usage.}
    \label{fig:cpu-baseline}
\end{figure}

Figure \ref{fig:cpu-baseline} presents the CPU utilization in \starlit. In the figure, we used 5 CPUs and did not use any CPU-specific library to run Paillier encryption, used as a subroutine in SecureBoost.

\subsection{Applications of Starlit's Components}\label{sec::components-applications}

Through \starlit's development, we demonstrated privacy preserving creation of features that are a function of data held by multiple parties (e.g. the string matching features on users' features held by \aut and the clients). This flexible feature creation is an essential building block of any federated learning problem. 

We have additionally developed a \textit{general-purpose} framework that can be used (in other contexts) for selecting optimal obfuscation mechanisms to safeguard flags (any categorical variables), aiming to maximize inference privacy while adhering to specified constraints on utility and local differential privacy. This framework offers flexibility in expressing utility constraints and can be adapted to prioritize utility maximization under specified constraints on inference privacy and guarantees of local differential privacy. Our modular design and integration between FATE and \flower enable not just SecureBoost, but also a broad range of FATE functions to be executed via \flower. Furthermore, the \flower adapter architecture we have built contributes to extending the use of \flower to vertically partitioned use cases.

\clearpage

\end{document}